\documentclass[10pt,journal,compsoc]{IEEEtran}

\ifCLASSOPTIONcompsoc
\usepackage[nocompress]{cite}
\else
\usepackage{cite}
\fi
\ifCLASSINFOpdf
\else
\fi
\usepackage{gensymb}

\usepackage{amsmath,amssymb,amsfonts}
\usepackage{graphicx}
\usepackage{textcomp}
\usepackage{xcolor}
\usepackage{tikz}
\usetikzlibrary{bayesnet}

\usepackage{mathrsfs}
\usepackage[colorlinks,linkcolor=red,citecolor=blue]{hyperref}
\usepackage{amsmath}
\usepackage{amsfonts,subfigure}
\usepackage{amssymb}
\usepackage{multirow}
\usepackage{booktabs}
\usepackage[ruled]{algorithm2e}

\usepackage{graphicx}
\usepackage{amsthm}

\graphicspath{{figure/}}

\newtheorem{definition}{Definition}[section]

\usepackage{bm}
\usepackage{color,soul}

\usepackage{tikz}

\usepackage{multirow}
\usepackage{adjustbox}
\usepackage{environ}
\usepackage{setspace} 

\newenvironment{alignSmall}{\nobreak\small\noindent\align}{\endalign}
\newenvironment{alignFootnotesize}{\nobreak\footnotesize\noindent\align}{\endalign}
\newenvironment{alignScriptsize}{\nobreak\scriptsize\noindent\align}{\endalign}

\NewEnviron{myAlignS}{%
	\begin{singlespace}
		\vspace{-3pt}
		\begin{alignSmall}
			\BODY
		\end{alignSmall}
		\vspace{-5pt}
	\end{singlespace}
}

\NewEnviron{myAlignSS}{%
	\begin{singlespace}
		\vspace{-3pt}
		\begin{alignFootnotesize}
			\BODY
		\end{alignFootnotesize}
		\vspace{-5pt}
	\end{singlespace}
}

\NewEnviron{myAlignSSS}{%
	\begin{singlespace}
		\vspace{-5pt}
		\begin{alignScriptsize}
			\BODY
		\end{alignScriptsize}
		\vspace{-5pt}
	\end{singlespace}
}

\usepackage{enumitem}

\newcommand{\eqnref}[1]{Eq. (\ref{#1})}
\newcommand{\defref}[1]{Def. \ref{#1}}
\newcommand{\secref}[1]{Sec. \ref{#1}}
\newcommand{\tableref}[1]{Table~\ref{#1}}
  
\newcommand{\figref}[1]{Fig. \ref{#1}} 
\graphicspath{{image/}}
\begin{document}

\title{Multi-task Weakly Supervised Learning for Origin–Destination Travel Time Estimation
}

%
%
%
%


\author{Hongjun~Wang,
	Zhiwen~Zhang,
	Zipei~Fan, 
	Jiyuan~Chen,\\ 
        Lingyu~Zhang,
	Ryosuke	Shibasaki
	and
    Xuan Song
	\IEEEcompsocitemizethanks{
		\IEEEcompsocthanksitem Hongjun Wang, Jiyuan Chen  and Xuan Song are with (1) SUSTech-UTokyo Joint Research Center on Super Smart City, Department of Computer Science and Engineering
        (2) Research Institute of Trustworthy Autonomous Systems, Southern University of Science and Technology (SUSTech), Shenzhen, China.
		E-mail: {wanghj2020,11811810}@mail.sustech.edu.cn and songx@sustech.edu.cn. \hfil\break 
		\IEEEcompsocthanksitem Zipei Fan, Ryosuke Shibasaki and Zhiwen Zhang are The University
		of Tokyo, 5-1-5 Kashiwanoha, Kashiwa-shi, Chiba, 277-8561, Japan; emails: zhangzhiwen@csis.u-tokyo.ac.jp, fanzipei@iis.u-tokyo.ac.jp, and shiba@skl.iis.u-tokyo.ac.jp\hfil\break 
          \IEEEcompsocthanksitem Lingyu Zhang is Research Institute of Trustworthy Autonomous Systems, Southern University of Science and Technology (SUSTech), Shenzhen, China. emails: zhanglingyu@didiglobal.com \hfil\break 
		\IEEEcompsocthanksitem Corresponding to 	Zipei Fan, Xuan Song;
		\IEEEcompsocthanksitem Hongjun Wang, Zhiwen Zhang equal contribution;
	}
}

%
%

\markboth{Journal of \LaTeX\ Class Files,~Vol.~XX, No.~X, August~201X}%
{Shell \MakeLowercase{\textit{et al.}}: Bare Demo of IEEEtran.cls for Computer Society Journals}
%

\IEEEtitleabstractindextext{%
\begin{abstract}
Travel time estimation from GPS trips is of great importance to order duration, ridesharing, taxi dispatching, etc. However, the dense trajectory is not always available due to the limitation of data privacy and acquisition, while the
origin-destination (OD) type of data, such as NYC  taxi data,  NYC  bike data, and  Capital Bikeshare data,  is more accessible. 
To address this issue, this paper starts to estimate the OD trips travel time combined with the road network. Subsequently,  a \underline{M}ulti-task \underline{W}eakly \underline{S}upervised \underline{L}earning Framework for \underline{T}ravel \underline{T}ime \underline{E}stimation (MWSL-TTE) has been proposed to infer transition probability  between roads segments, and  the travel time on road segments and intersection simultaneously. 
Technically, given an OD pair, the transition probability intends to recover the most possible route. And then, the output of travel time is equal to the summation of all segments' and intersections' travel time in this route. A novel route recovery function has been proposed to iteratively maximize the current routes' co-occurrence probability, and minimize the discrepancy between routes' probability distribution and the inverse distribution of routes' estimation loss. Moreover, the expected log-likelihood function based on a weakly-supervised framework has been deployed in optimizing the travel time from road segments and intersections concurrently. We conduct experiments on a wide range of real-world taxi datasets in Xi’an and Chengdu and demonstrate our method's effectiveness on route recovery and travel time estimation. 

\end{abstract}
	
	\begin{IEEEkeywords}
		Travel Time Estimation, Urban Computing, Weakly Supervised Learning
\end{IEEEkeywords}}

\maketitle

\section{INTRODUCTION}

With the emergence of newly-developed applications, estimating travel time has become one of the hottest topics, which is of great importance to route planning, taxi dispatching, and ride-sharing in recent years. 
In the early phase, the data of real traffic state is mainly collected from loop sensors, which can only provide the individual travel time in a certain road segment and usually face the sparse issue. Recently, an alternative solution  is to use floating car data. The floating cars equipped with GPS receivers, including taxis, buses, private cars, and online ride-hailing, record time stamps, longitude, latitude, speed, and other information at regular intervals, which can reflect the vehicle's operation status. 

\begin{figure}[t]
	\centering
	\includegraphics[width=1\linewidth]{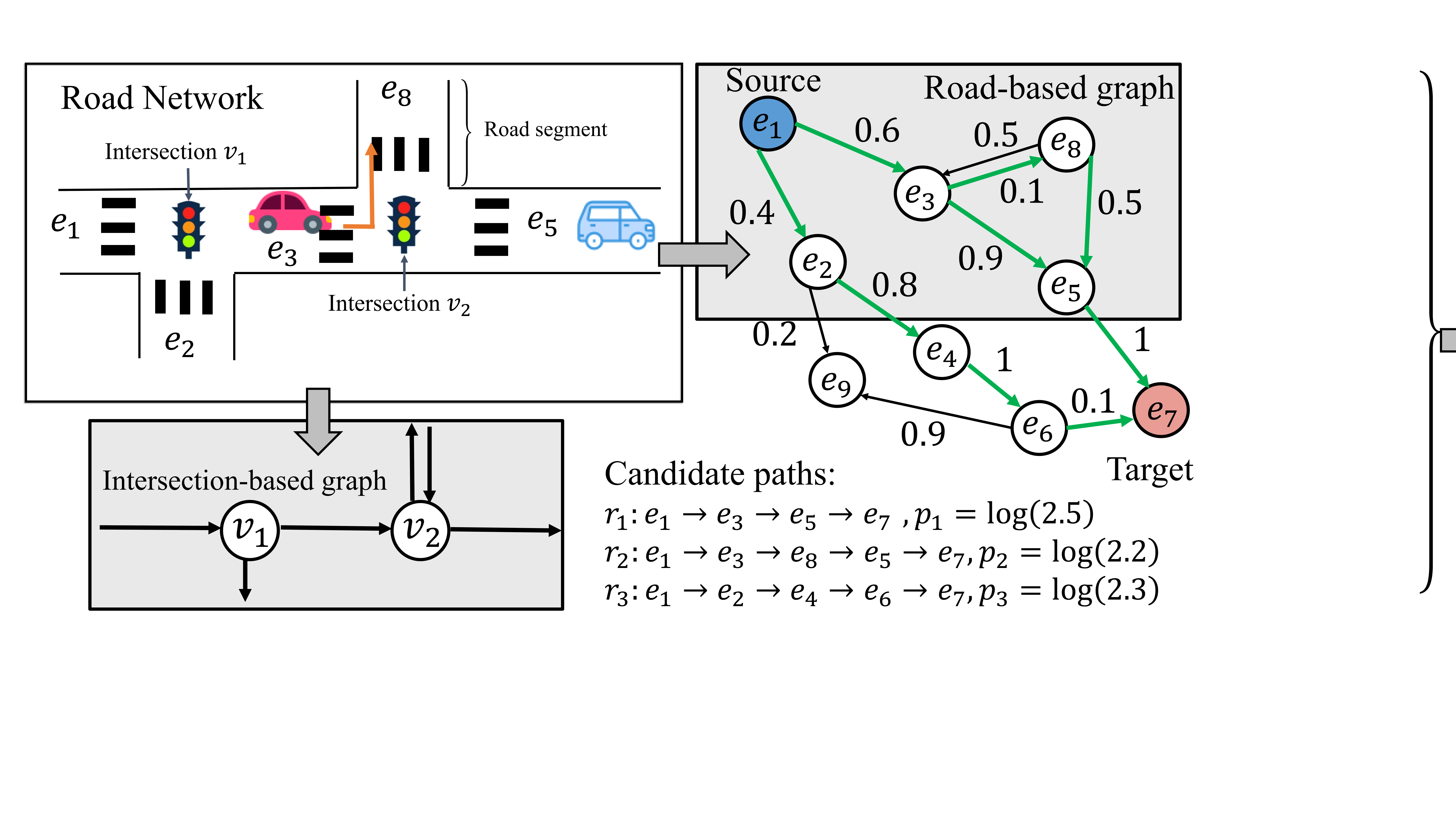}
	\caption{The motivation in this paper is illustrated above. Given any OD pair, we want to recover the path by searching the  maximum route probability learned by the model, and construct a dual graph from road-based and intersection-based aspects to estimate the  travel time of road segment and intersection respectively. }
	\label{fig:motivation}
\end{figure}

As a result, a good deal of travel time estimation techniques based on floating car data have been proposed in different scenes, such as dense trajectory \cite{wang2018learning,wang2018will,li2019learning}, low-sampling-rate trajectory \cite{wu2016probabilistic,wang2014travel,sanaullah2016developing}. However, due to the  privacy concern and data acquisition problems, extensive works focus on inferring the travel time from OD data, \textcolor{black}{which only gives} the origin-destination location, such as  finding nearby neighbors \cite{wang2019simple},
such as distance-based \cite{jindal2017unified} or representation-based \cite{li2018multi} neighbors.
In general, the OD type of data is more available than the dense type, and multiple sources of OD data have been released, for example,  the NYC taxi data\footnote{https://www1.nyc.gov/site/tlc/about/tlc-trip-record-data.page}, NYC bike data\footnote{https://data.cityofnewyork.us/Transportation/Bike-Data/374u-5ie7} and Capital Bikeshare Data\footnote{https://www.capitalbikeshare.com/system-data}. However, as far as we know, previous literature omits the factor of the road network, which often leads to a high estimating error. Since the total travel time of the trajectory is equal to the sum of the travel time of all road segments and intersections (e.g., waiting traffic signal).  Each traffic condition in road segment change will affect the total travel time.  With the road network introduced, here we face three intractable problems:
\begin{itemize}
    \item[1)] \textit{How to recover the route when only OD pairs are given}.
    \item[2)] \textit{How to effectively estimate the travel time when the route has been obtained.}
    \item[3)] \textit{How to learn  features from complex road network.}
\end{itemize}
At first glance, given a pair of origin-destination, the shortest path algorithm (e.g., Dijkstra's algorithm) is a natural choice for problem 1) because people usually choose paths that are similar to the
shortest path with less number of turns. However, the shortest path in the geometry aspect may not always match the definition of the 'shortest path' in the driver's route choice. For example, some resident or tertiary types of road are shorter than primary and trunk types of road, but they are more vulnerable to congestion, since the complex traffic state (many pedestrians), or narrowness of road width. How to encode the road features into the road search procedure? One way can be done by learning the transition probability between road segments and inferring the route via the search for the maximum route probability, where the superiority of this approach is that the character of the road will be considered in every search process.  
Inspired by \cite{cordonnier2019extrapolating}, this article employs the graph neural network (GCN) to learn the features, such as road type, road length, road sign, and road lanes,  of each road segment. Consequently, the problem of ignoring natural road networks in the shortest-path algorithm can be alleviated. \figref{fig:motivation} shows an example of searching candidate path through transition probability. Based on the Markov assumption, the routing probability can be obtained by multiplying the probabilities and equal to the sum of the log probabilities. The candidate paths $r_1,r_2,r_3$ are acquired by Depth  First Searching  (DFS) algorithm with pruning operation. 

As we mentioned, this paper infers the overall travel time of a given route by summing up the travel time of all the road segments and intersections on that route. This raises the question of how to estimate the travel time of road segments and  intersections reasonably. 
One concern is how to model the differences in individual driving behaviors, since given a specific OD pair, the travel time in the same time interval is varied. 
To address this problem, we here model the travel time with uncertainty, which means that each road/intersection travel time follows  certain distribution (e.g., Lognormal). Those roads/intersections tend to be provided a large variance $\sigma$, for example, with large crowd flow. In conclusion, the uncertain travel time of route generated by route searching has been obtained. To effectively optimize the distribution, this paper formulates it as the inexact supervision learning  problem \cite{zhou2018brief}.
One of the most well-known examples of inexact supervision is the drug activity prediction problem \cite{dietterich1997solving}, which predicts if a molecule induces a given effect. Inexact supervision deals with training data arranged in sets called \textit{bags}, and the labels are only annotated on  \textit{bags}. For modeling the uncertainty, given a pair of origin and destination, we consider the travel time label annotated on the unobserved routes (\textit{bags}) and use the normal
distribution \cite{richardson1978travel,rakha2006estimating,arezoumandi2011estimation} to model the uncertainty for each road segment and the interaction travel time in \textit{bags}. We drive the objective function (\eqnref{equ:obj1}) based on the assumption of aggregated observation and Markov chain, and solve it with a general inexact learning probabilistic framework \cite{zhang2020learning} and an iterative route recovery algorithm. 


After solution 1) and 2) has been discussed, we finally introduce how to learn the meaningful features from a complicated road network, since multiple factors will affect the traffic condition, such as road types, road lanes, speed limitations, and traffic signals.  To learn the intricate relation of road networks, many existing works modeling the problem of estimating travel time from either road segments \cite{li2019learning,wang2018learning} or intersections \cite{wang2019learning}, but do not assemble those features simultaneously. However, we argue that this will cause error accumulation with the road segments increasing. To fill this gap, in this work, we construct a dual graph comprised of the road-based graph and the intersection-based graph, and estimate the travel time by summing up all road segments and intersections on one route. In the meanwhile, we also take the connected relation\footnote{https://wiki.openstreetmap.org/wiki/Key:highway} into consideration from one road segment to one road segment, such as \textit{primary $\rightarrow$ secondary}, or \textit{trunk $\rightarrow$ residential}, and one intersection to one intersection, such as \textit{tertiary} and \textit{residential}. We introduce the Relational Graph Convolutional Networks (R-GCNs) \cite{schlichtkrull2018modeling} to learn the complex connected relation of the road network. Moreover, to solve the problem of losing local patterns when expanding the receptive neighborhood in GCN, we combine the Relational Graph Convolutional Networks, and gated recurrent unit (GRU) \cite{chung2014empirical} together as the stacked architecture to capture both global and local features \cite{wang2020traffic}. \figref{fig:motivation} represents the procedure to construct the dual graph (gray box) of the road network and to recover the route from the candidate set that is searched from the transition probability.



The main contributions of this paper can be summarized as follows.

\begin{itemize}
	\item[$\bullet$] We propose a multi-task framework to estimate the travel time of road segments and intersection, and transition probability simultaneously. To the best of our knowledge, it is the first attempt to recover the route and jointly model the factor of intersection and road segments in the general OD travel time estimation problem. 
	\item[$\bullet$] For the first time, we consider the estimation of the OD travel time as the weakly supervised learning problem, since the observation of the OD travel time is annotated with a bag of unobserved routes. This paper aims to infer each road segment and the intersection travel time distribution from the aggregation observation.
	\item[$\bullet$] We validate the effectiveness of travel time estimation and route recovery using large-scale datasets from the real world in Chengdu and Xi’an, respectively, which significantly outperform current methods.
\end{itemize}

Here, we list the organization in this paper: \secref{sec:related} gives the related works, including weakly supervised learning, travel time estimation, as well as route estimation. \secref{sec:prelim} introduces the preliminary knowledge, such as the road network, the origin-destination. \secref{sec:Formulation} provides the definition of our formulation, assumptions, and objective function. \secref{sec:Methodology} gives the methodology of our MWSL-TTE. \secref{sec:exps} and \secref{sec:case_study} conduct the qualitative  and  quantitative experiments respectively to demonstrate the superiority of MWSL-TTE. \secref{sec:concluds} gives a summary of this paper and future work.

\section{RELATED WORK}\label{sec:related}

In this section, we will discuss several relevant topics about weakly supervised learning, travel time estimation, as well as route estimation.
\subsection{Weakly Supervised Learning}
Since the general supervised learning method requires each data in the training set to be labeled, this expensive labeling consumes a lot of manpower and time. Therefore, learning under the condition of weakly supervised information has become a hot research topic in the field of machine learning in recent years \cite{zhang2021weakly,li2019towards,nodet2021weakly}.
The weakly supervised learning methods focus on addressing the low-quality labels scenarios 1) incomplete supervision \cite{settles2009active} : only part of data can be labeled. 2) inexact supervision \cite{dietterich1997solving}: only have coarse-grained labels . 3) inaccurate supervision \cite{frenay2013classification} : only part of the data owns true labels. The task of estimating travel time from OD can be considered as the  inexact supervision belonging to the category of Weakly Supervised Learning, where the only observations are the total travel time and OD locations, but the accurate routes are unknown. Different from traditional approaches \cite{wang2019simple,tiesyte2008similarity} searching similar historical trajectories from data but ignoring the city road network structure, in this paper, we aim to infer the potential route from OD by using transition probability between road segments, where a set of potential routes can be seen as a \textit{bag}, and the OD travel time can be obtained by summing up the estimated times of the road
segments in \textit{bag}. To the best of our knowledge, we are the first to introduce the problem of travel time estimation into the inexact supervision framework.

\subsection{Travel Time Estimation}
\textcolor{black}{Various TTE implementations were classified into three groups, traditional approaches, deep learning-based approaches and graph neural network-based approaches. Traditional approaches for TTE include the road-segment-based and path-based methods. The road segment-based methods \cite{wu2004travel,sevlian2010travel} coarsely forecast the route travel time by summing up all estimated times of roads by using the data collected from sensors like magnetometer detectors or highway cameras, which omitted the necessary factor of intersection and relationship among road segments. And the path-based methods address the above challenges mainly by  nearest neighbors search \cite{wang2019simple,tiesyte2008similarity} and trajectory regression  \cite{luo2013finding,sevlian2010travel,yang2018pace}. Nearest neighbors search (NNS) finds nearby historical trajectories according to the assumption that the routes with similar origins and destinations own close travel time. Trajectory regression methods predict the whole route travel time based on the given historical trajectories. }

\textcolor{black}{Recently, deep learning-based approaches have become especially important in the task of TTE. These approaches can be divided into two groups, classical deep learning-based methods and graph deep learning-based methods. Some classical methods such as deep neural networks (DNNs) \cite{wang2018learning} and convolutional neural networks (CNNs) \cite{fu2019deepist,wang2018will}  have been successfully applied in TTE. For example, Deep-TTE \cite{wang2018learning} proposed a CNN-based framework to integrate various types of attribute information (such as weather, time ID and driver ID) for TTE. However, most of these methods model the road network as a grid-based map, but they ignore the graphical structure of real-world road network.}

\textcolor{black}{To fully utilize spatial information, GNN is an emerging tool to analyze the topological relations of graph-structured traffic data. Especially, Spatial-Temporal Graph Neural Network (STGNN) \cite{yu2017spatio} is a framework that integrates GNN and temporal processing modules, which can handle spatial relations and temporal trends simultaneously. Due to the spatio-temporal characteristics of the real-world road network, STGNN are widely adopted in TTE. For example, diffusion convolutional recurrent neural network (DCRNN) \cite{li2017diffusion} modeled the graph-structured traffic data as a diffusion process on a directed graph and transformed spatio-temporal features into a seq2seq framework. 
ASTGNN \cite{guo2021learning} proposes a trend-aware multi-head attention mechanism to capture multiple potential correlations in traffic forecasting. However, these works only consider the spatio-temporal attributes of road segments but ignore the interactive correlations between intersections and road segments. Meanwhile, both the real route of OD pairs and road condition also have an important influence on TTE.}

%
\subsection{Route Estimation}
Another bunch of research in travel time estimation is to solve the issue of sparse trajectory due to the privacy,
business competition \cite{wu2016probabilistic,wang2014travel,sanaullah2016developing}, and  limitation of GPS devices, or in the scene of  ETC \cite{yang2018sharededge,chen2018dyetc} and surveillance cameras \cite{shao2020estimation}. A common strategy for solving the sparse trajectory is to infer the potential route based on the information of the road network. Reference \cite{wu2016probabilistic}  applies the inverse reinforcement learning to  learn the latent cost
(reward) of a road through historical data, and proposed  Exact Route Search approach to find the maximum probabilistic route based on dynamic programming. However, route search-based algorithms are only adapted in low-sampling rate trajectories, but not  OD problem, due to the heavy computational cost.   Because of  the large distance between a pair of toll stations or surveillance cameras, a frequently used path inferring algorithm is based on the Depth First Searching algorithm to find all possible \textit{simple paths} that the one road segment can only appear at most once. However, those methods omitting the real traffic condition tend to generate the unreal route in the path inferring procedure. To this end, in this paper, we combine the transition probability and route search approach together to find the optimal route based on the real travel time and road network structure.  

%
%

\section{Preliminaries}\label{sec:prelim}

We start with giving the definition about the road network, Origin-Destination, simple path as well as route.

\begin{definition}
	\textbf{Road Network}. A road network is a directed graph $\mathcal{G}=(\mathcal{V},\mathcal{E},\pi)$, where $\mathcal{V}$ denotes the set of nodes, $\mathcal{E} \subseteq \mathcal{V}\times \mathcal{V} $ is the set of directed edges, and $\pi$ is a node's feature set. This paper uses $\mathcal{G}_v$ and $\mathcal{G}_e$ to denote the node-wise and link-wise graphs respectively. For the $\pi$ in $\mathcal{G}_v$, 
    the features of $\mathcal{V}$ can be such as, junction types, and traffic signals.  For the \textcolor{blue}{$\pi$} in $\mathcal{G}_e$,  the features of $v$ can be such as, road types, road lanes, and speed limits. 
\end{definition}

\begin{definition}
	\textbf{Origin-Destination (OD)}. In this paper, a OD pair  represents a tuple with  $\{e_i,e_j,t^{th}\}$, where $e_i$ and $e_j$ denote the start and end road segment, respectively, and  $t^{th}$ is the start time interval of a day (e.g.,  8:00am-9:00am). Note that we assume the traffic conditions for all road segments and intersections  are invariable within the same time slot.
\end{definition}

\begin{definition}
	\textbf{Simple Path}. A simple path $t_r$ can be presented as a series of time-ordered  links. We have $t_r: e_{1} \rightarrow e_{2} \rightarrow \cdots \rightarrow e_{|t_r|}$, where each link satisfied $e_{i} \neq e_{j}$.
\end{definition}

\begin{definition}\label{def:route}
	\textbf{Route}. A route $r$  in this paper is a sequence alternating with links and intersections. We have $r: e_{1} \rightarrow v_2 \rightarrow e_{3} \rightarrow \cdots  e_{K-1}\rightarrow  v_{K}$, where $v_i$ is the intersection of edges pair $(e_{i-1},e_{i})$, and $K$ is the length of $r$.
\end{definition}

\section{Problem Formulation}\label{sec:Formulation}

\begin{figure}[t]
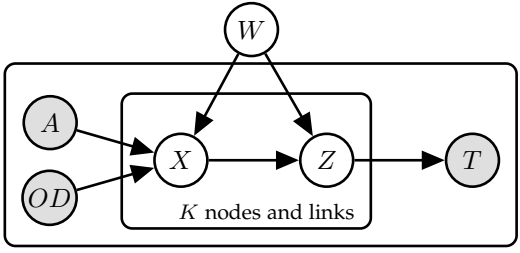

	\centering
	\tikz[line width=1pt]{
		\node[latent] (x) {$X$};%
		\node[obs, left=of x,yshift= 0.5 cm] (pi) {$A$};%
		\node[obs, left=of x,yshift= -0.5 cm] (od) {$OD$};%
		\node[latent, right = of x, xshift=0.2cm] (z) {$Z$};%
		\node[obs, right = of z, xshift=0.2cm,yshift=-0 cm] (t) {$T$};%
		\node[latent, above = 10mm of z, xshift=-1.0cm] (w) {$W$};%
		\edge {w,x} {z}
		\edge {w} {x}
		\edge {z} {t}
		\edge {od,pi} {x}
		\plate [inner sep=3.0mm, xshift=-0.10cm, yshift=0.2cm] {plate1} {(x)(z)} { $K$ nodes and links}; %
		\plate [inner sep=2.0mm, xshift=-0cm,yshift=0.2cm] {plate1} {(pi)(x)(z)(t)(od)} { }; %
	}
	\caption{ The graphical model of the data generating process. 
	}\label{fig:data_gens}
\end{figure}

To overcome the previous issue with ignoring the road network in the OD-TTE problem, we here intend to give a formulation under the weakly supervised learning. This solution is motivated by the advance in weakly supervised learning and GCN path inference.
\begin{figure*}[t]
	\centering
	\setlength{\belowcaptionskip}{-0.1cm}
	\includegraphics[width=1\linewidth]{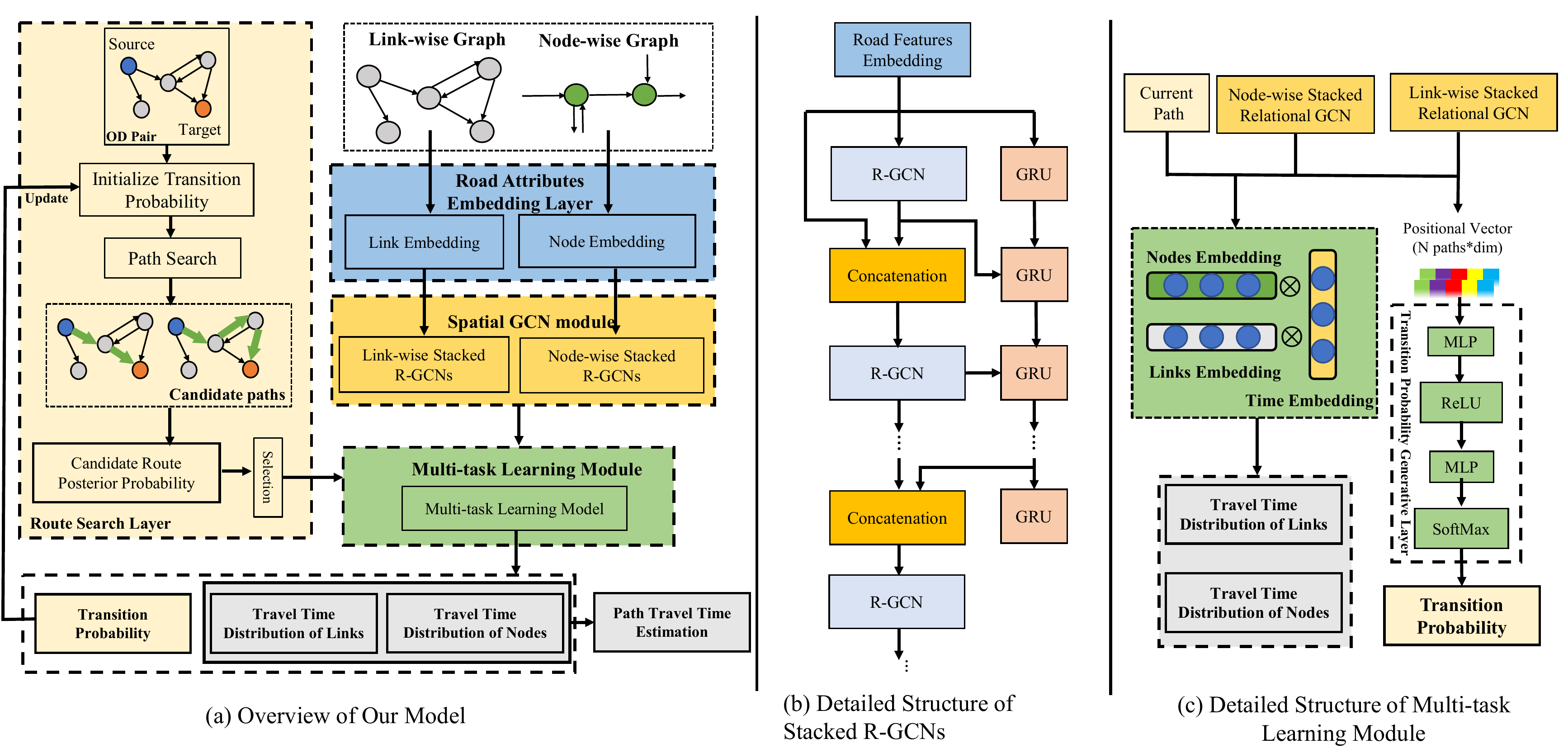}
	\caption{\textcolor{black}{Framework design of MWSL-TTE. The overview of our proposed model is depicted in (a), which includes the route search layer, road attributes embedding layer, spatial GCN module, and multi-task learning module. (b) is the architecture of the stacked R-GCNs layer, which attempt to capture both global and local relation by modeling the GRU and R-GCNs together. (c) is our multi-task learning layer, which will estimate the travel time of links, intersections, and transition probability simultaneously. The transition probability will be updated when the multi-task learning layer is output, and then, the path search algorithm will work to find the candidate paths.}}
	\label{fig:framework}
\end{figure*}

Given a pair of origin and destination, estimating the travel time is to infer the total time cost. Traditionally, the formulation of TTE can be divided into two parts: 1) inferring the future traffic through historical state \cite{li2019learning,jin2021spatial}, 2) online infer traffic through real-time trajectories \cite{james2021citywide}. The previous one is mainly related to the robust traffic state, which is calculated on either dense trajectories \cite{li2019learning} or loop detectors \cite{yu2017spatio}. Another one intends to resolve the sparse issue by estimating (imputation) citywide-level traffic state from fewer real-time trajectories. Since this paper is under the  OD scenario and hard to give a valid historical traffic state, we here follow the online TTE formulation. Formally, this paper follows the assumption that the traffic state at one road segment in a specific time interval $\Delta \mathbf{t}=60mins$, e.g., 10:00 am-11:00 am, under the same distribution, such as, Gaussian distribution $\mathcal{N}(\mu,\sigma)$ with $\mu=60s$ and $\sigma=1$. Subsequently, suppose the current time  is $\mathbf{t}$, we train the real-time OD pairs in time slot $[\mathbf{t},\mathbf{t}+\Delta \mathbf{t}_1)$, the online traffic state will be completed and evaluated  with the OD pairs in $[\mathbf{t}+\Delta \mathbf{t}_1,\mathbf{t}+\Delta \mathbf{t})$.

\textcolor{black}{About the training procedure of MWSL-TTE, \figref{fig:data_gens} illustrates the data generating process with graphical representation. $OD$ are the features of origin-destination locations. $X_{1:K}$\footnote{The subscript for example $X_{1:K}$ denotes an abbreviation for the set $\{X_1,X_2,\cdots,X_K\}$} stands for the features vectors of unobserved $K$ nodes and links in route $r: e_{1} \rightarrow v_2 \rightarrow e_{3} \rightarrow \cdots  e_{K-1}\rightarrow  v_{K}$, and $Z_{1:K}$ is the  unobservable \textit{travel time} of nodes and links. We assume that $r$ is conditioned on $OD$ and  transition matrix $A$, where $A_{i,j}$ indicates the transition probability from $e_i$ to $e_j$. 
Therefore, we have $p(r|A;OD)=p(X_{1:K}|A;OD)$, where
$A$ is generated by multi-task function $f$ with $A=f(W_A;\mathcal{G}_e)$,  $\mathcal{G}_e$ is the link graph, and $W$ is the learnable parameters. Meanwhile, $Z$ also  under a parametric distribution $p(Z_{1:K}|\theta_z=f(X_{1:K}; W_Z;))=p(Z_{1:K}|(\mu,\sigma)=f(X_{1:K}; W_Z;))$ on the factors of $X_{1:K}$ and $W_Z$, where $\mu$ and $\sigma$ are the mean and variance of Gaussian distribution.  For the relation between $Z_{1:K}$ and aggregate observation of travel time $T$, we have the following definition: }
\begin{definition}\label{def:tte}
Given a route $r_i: e_{1} \rightarrow v_1 \rightarrow e_{2} \rightarrow \cdots v_{|r_i|-1} \rightarrow e_{|r_i|}$ under a pair of $OD$,  and  it’s unobserved travel time $Z_{1:K}=\{\tilde{t}_{e_{1}},\tilde{t}_{v_2},\tilde{t}_{e_3}, \allowbreak \cdots \allowbreak \tilde{t}_{v_{K-1}}, \tilde{t}_{e_{K}}\}$. The aggregate functions $Q$ can be defined as 
\begin{align}
Q(Z)=\tilde{t}_{e_{1}}+\tilde{t}_{v_1}+\tilde{t}_{e_2}\cdots +\tilde{t}_{v_{K-1}}+ \tilde{t}_{K},
\end{align}
where an aggregate function $Q: Z\rightarrow T$ is a mapping function from unobserved variable $Z$ to observation $T$. Since we assume $Z$ follows a Gaussian distribution, we can written $T=Q(Z)=\sum_{i}\tilde{t}_i=\sum_{i}\mu_i$ according to the nature of additivity: $X+Y \sim N\left(\mu_{1}+\mu_{2}, \sigma_{1}^{2}+\sigma_{2}^{2}\right)$, where $X \sim N\left(\mu_{1}, \sigma_{1}^{2}\right) Y \sim N\left(\mu_{2}, \sigma_{2}^{2}\right)$.
\end{definition}
Subsequently, we summarize our assumptions below.

\noindent\textbf{Assumption 1}(Aggregate observation assumption) $p(T \mid X_{1:K}, \allowbreak Z_{1:K})  = p(T \mid  Z_{1:K})$

We here assume that the observation $T$ is conditionally independent $X_{1:K}$ when given $Z_{1:K}$ (\defref{def:tte}).  Intuitively, given $Z$, in fact, the travel time $T$ can be obtained by summing up all components in $Z$. 

\begin{table}[t]
	\centering
	\renewcommand\arraystretch{1.0}
	\caption{\textcolor{black}{Partial Symbols Description.}}
	\label{tab:symbol}
	
	\begin{tabular}{c|l}
		\toprule
		Notation & Description\\
		\hline
		$A$ & Links inner transition probability matrix\\
		$Q$ & Aggregate function\\
		$\Omega_{OD}$ &  Set of candidate paths from $O$ to $D$\\
		$f$ & Multi-task function\\
		$X$ & Features of links and nodes\\
		$Z$ & Unobservable travel time of links and nodes\\
		$\mathcal{G}_v$ and $\mathcal{G}_e$ & Node-wise and link-wise graph respectively\\
		$T$ and $\widetilde{T}$ & Ground truth  and estimated  of travel time\\	
		$\mathcal{T}$ & Time embedding vector\\	
		$W$ & Learnable parameters\\	
		\bottomrule
	\end{tabular}
\end{table}

\noindent\textbf{Assumption 2} (Markov chain assumption)\label{amp:markov}
$p(Z_{1: K} \mid X_{1: K})=p(Z_{1} \mid X_{1})\prod_{i=2}^{K}    p(Z_{i+1}  \allowbreak \mid X_{i}; Z_{i})$

We assume that the road travel times $Z_{i+1}$ are mutually independent except for $Z_{i}$, which is consistently under the assumption of Markov chain  and extensive applications in trajectory data mining \cite{brakatsoulas2005map,li2015inferring}.
Furthermore, since $T$ can be determined by function $Q$, the conditional probability for $p(T\mid Z_{1: K})=\delta_{Q\left(Z_{1: K}\right)}(T)$, where $\delta(\cdot)$ represents the \textit{Dirac delta function}.

\textcolor{black}{To sum up, the objective function in this paper can be defined as 
\begin{align}\label{equ:obj1}
&\max_{ X_{1: K}} ~ p\left(T ; X_{1: K}\mid A; OD\right) \nonumber \\ 
= &p\left(T \mid X_{1: K}\right) p(X_{1: K}|A; OD) \nonumber \\ 
=&\int_{Z^{K}}p\left(T; Z_{1: K} \mid X_{1: K}\right)d_{ Z_{1: K}} \ p(X_{1: K}|A; OD)\nonumber \\
=&\int_{Z^{K}} \delta_{Q\left(z_{1: K}\right)}(T) p(Z_{1: K} \mid X_{1: K})d_{ Z_{1: K}} \ p(X_{1: K}|A; OD) \nonumber \\
=&\underset{Z_{i} \sim p\left(Z_{i} \mid X_{i-1};Z_{i-1}\right)}{\mathbb{E}}\left[\delta_{Q\left(Z_{1: K}\right)}(T) \right]p(X_{1: K}|A; OD)
\end{align} 
Therefore, according to \eqnref{equ:obj1}, our training procedure could be split into two stages: 1) maximizes the  posterior probability $p(X_{1: K}|A; OD)$. 2) maximizes the conditional probability $p(Z_{1: K} \mid X_{1: K})$ to estimate each road segments and intersection travel time, which can be optimized by \textit{ expected log-likelihood} \cite{zhang2020learning}: 
\begin{align}
	\ell(W)=\mathbb{E}\left[\log p\left(T \mid X_{1: K} ; W\right)\right]\label{equ:obj2}
\end{align}
For ease of reference, some important notations are summarized in \tableref{tab:symbol}.}

\textcolor{black}{We here give a brief summarization of our problem formulation. In this paper, we aim to solve the two challenges in OD travel time estimation, which are uncertain routes and  uncertain travel time.  We intend to infer the potential route $r$ between source and destination by transition probability $A$. Subsequently, we optimize the travel time distribution $Z_{1:K}=\{\tilde{t}_{e_{1}},\tilde{t}_{v_2},\tilde{t}_{e_3}, \allowbreak \cdots \allowbreak \tilde{t}_{v_{K-1}}, \tilde{t}_{e_{K}}\}$ at $r$ via weakly supervised learning (\eqnref{equ:obj2}). We assume $\tilde{t}$ under Gaussian distribution $\mathcal{N}(\mu,\sigma)$. Therefore, transition probability $A$ and parameter at Gaussian distribution can be generated by $(A,\mu,\sigma)=f(W; \mathcal{G}_v; \mathcal{G}_e)$, where $\mu,\sigma$ are the mean and variance in Gaussian distribution, respectively.}
\section{METHODOLOGY}\label{sec:Methodology}

In this section, the MWSL-TTE will be detailedly introduced. Specifically, the overview of MWSL-TTE has been depicted in \figref{fig:framework} (a)
including with four components included road attributes embedding layer, spatial GCN module, route search layer, and multi-task learning module. \figref{fig:framework} (b) shows the inner structure of  the stacked R-GCNs layer, and \figref{fig:framework} (c) represents the multi-task learning module.

 \begin{figure}[t]
	\centering
	\setlength{\belowcaptionskip}{-0.5cm}
	\includegraphics[width=1\linewidth]{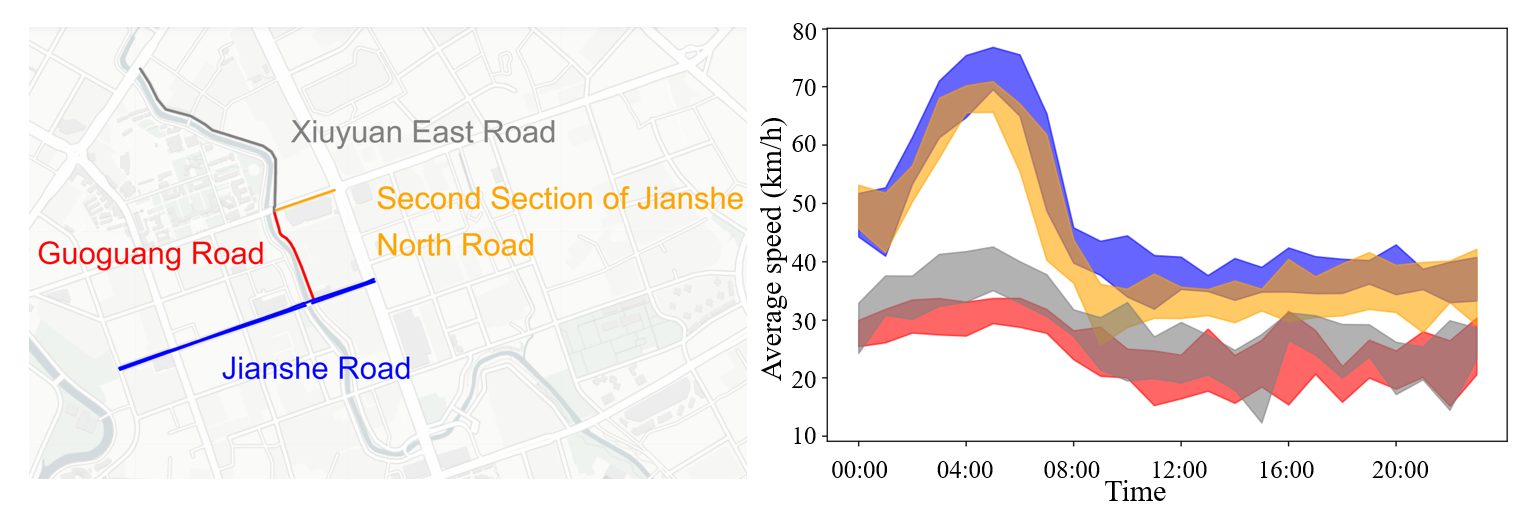}
	\caption{\textcolor{black}{In the speed distributions of neighbor road segments during the 7 days of National Day, we observe that the speed distribution is highly consistent with the connected type. The color indicate the road correspondingly.} }\label{fig:example}
\end{figure}

\subsection{Road Attributes Embedding layer}
\textcolor{black}{Let  each latent variable $\tilde{t}_i \in Z$ belong to Gaussian distribution $\mathcal{N}(\mu_i,\sigma_i)$, which is a  common assumption and widely be used in modeling the travel time distribution \cite{richardson1978travel,rakha2006estimating,arezoumandi2011estimation}. Given a pair of OD,  we have the conditional probability \textcolor{black}{$p(Z_{1:K}|\theta_z=f(X_{1:K}; W_Z))=p(Z_{1:K}|(\mu,\sigma)=f(X_{1:K}; W_Z))$}. Therefore, one of the \textcolor{black}{tasks} for neural network $f$ is to estimate the distribution parameters $\mu, \sigma$ for each road segment and intersection.  \textcolor{black}{Since the road features $X_{1:K}$  could affect the travel time estimation $Z$, we consider the follow statistical spatial factors as the important matters for road segments:}}
\begin{itemize}[leftmargin=*]
	\item[$\bullet$] \textit{Road types}: e.g., primary, primary link, secondary, secondary link, tertiary, residential, service road, etc.;
	
	\item[$\bullet$] \textit{Number of lanes}: how many traffic lines in the road;
	
	\item[$\bullet$] \textit{Otherwise features}: e.g., road length, whether it is a one way or not, limiting velocity, unique ID;
\end{itemize}
 and intersections, such as 
\begin{itemize}[leftmargin=*]
	\item[$\bullet$] \textit{Node tags}: e.g., speed camera, traffic signal, crossing sign, turn circle, stop sign;
	
	\item[$\bullet$] \textit{Node street count}: e.g., T-juction  X-junction, and 5-way junction: 
	
	\item[$\bullet$] \textit{Otherwise features}: e.g., unique ID, GPS \textcolor{black}{coordinate}.
\end{itemize}

\textcolor{black}{To obtain the feature representations of both links and nodes, we use the embedding method \cite{gal2016theoretically} to transform each categorical attribute into a low-dimensional feature vector by multiplying the spatial feature embedding matrices $E\in R^{n^{(s)}\times d^{(s)}}$. Here, $n^{(s)}$ represents the number of possible values of the categorical features, and $d^{(s)}$ represents the embedding dimension. This allows us to share efficient information among different road segments or intersections, so that rarely traveled segments could be learned from those frequently traveled with similar semantic meaning. Besides the categorical road attributes, we concatenate the obtained embedded feature vectors together with other road attributes (e.g., road length and GPS coordinate). Based on the above feature representations of the dual graph (link-wise and node-wise), we can obtain the corresponding input for the subsequent spatial GCN module.}

\subsection{Spatial GCN Module}\label{sec:Spatial GCN}
After the embedding characteristics of the road attributes have been obtained, we next introduce the spatial GCN module serving as  modeling the complex spatial relations from the dual graph. The motivation for introducing R-GCNs in the travel-time estimation problem has been represented in \figref{fig:example}. We can observe that the travel speed is highly similar with the connected types. For specifically, even though the Second Section of JianShe North road is the neighbor of Xiuyuan East  and  Guoguang road, its speed distributions are more related to JianShe road, where their road types are the same. Therefore, we are concerned that  the features of \textit{road types}, the connected types, for example, \textit{resident} $\rightarrow$ \textit{secondary} (link level), and \textit{secondary} (node level), are also important. To this end, here we introduce the Relational Graph Convolutional Networks \cite{schlichtkrull2018modeling} in our model, which can be defined as 
\begin{align}
	\boldsymbol{h}_{i}^{(l+1)}=\sigma\left(\sum_{\boldsymbol{r} \in \mathcal{R}} \sum_{j \in \mathcal{N}_{i}^{r}} \frac{1}{c_{i, r}} W_{r}^{(l)} \boldsymbol{h}_{j}^{(l)}+W_{0}^{(l)} \boldsymbol{h}_{i}^{(l)}\right),
\end{align}
where $\boldsymbol{h}_{i}^{(l)} \in \mathbb{R}^{d^{(l)}} $ is the hidden state of road $r_i$ in the $l^{th}$ layer of model with dimensionality $d^{(l)}$. $\mathcal{N}_{i}^{r}$ denotes the set of neighboring road segment/intersection indices under the relation $\boldsymbol{r}$. $W_{r}^{(l)}, W_{0}^{(l)} $ are the learnable parameters, $c_{i, r}$ is the normalization constant, and $\sigma(\cdot)$ is the activation function. In this paper, we set $c_{i, r}=| \mathcal{N}_{i}^{r} |$. 

Next, we will introduce the stacking operation based on R-GCNs. The stacking operation has recently been demonstrated to prevent local information loss \cite{wang2020traffic,luan2019break}. \textcolor{black}{Thus, we model the temporal trends in the stacked GCN architecture combined with the spatial feature representations of both nodes and links. In this paper, we use a gated recurrent unit  (GRU) \cite{chung2014empirical} as a temporal processing module to incrementally concatenate multi-scale features, which can be written as} 

\textcolor{black}{
\begin{footnotesize}
	\begin{align}
		\boldsymbol{c}^{(0)}&=\operatorname{GRU}\left(\boldsymbol{h}^{(0)}, \boldsymbol{c}^{(-1)}\right),\nonumber \\
		\boldsymbol{h}^{(1)}&=\sigma\left(\sum_{r \in \mathcal{R}} \sum_{j \in \mathcal{N}_{i}^{r}} \frac{1}{c_{i, r}} W_{r}^{(0)} \boldsymbol{h}_{j}^{(l)}+W_{0}^{(0)} \boldsymbol{h}^{(0)}\right), \nonumber \\
		\boldsymbol{h}^{(2)}&=\sigma\left(\sum_{r \in \mathcal{R}} \sum_{j \in \mathcal{N}_{i}^{r}} \frac{1}{c_{i, r}} W_{r}^{(1)} \left[\boldsymbol{h}_{j}^{(0)}, \boldsymbol{h}_{j}^{(1)}\right]+W_{0}^{(1)} \left[\boldsymbol{h}^{(0)}, \boldsymbol{h}^{(1)}\right] \right), \nonumber \\
		\boldsymbol{h}^{(l+1)}&=\sigma\left(\sum_{r \in \mathcal{R}} \sum_{j \in \mathcal{N}_{i}^{r}} \frac{1}{c_{i, r}} W_{r}^{(l)} \left[\boldsymbol{c}_{j}^{(l-1)}, \boldsymbol{h}_{j}^{(l)}\right]+W_{0}^{(l)} \left[\boldsymbol{c}^{(l-1)}, \boldsymbol{h}^{(l)}\right] \right), \label{equ:stack_gcn}  \\
		l&=2,3, \ldots, n-1, \nonumber
	\end{align}
\end{footnotesize}
where $\boldsymbol{c}^{(l)}$ is the hidden state of the output of GRU, and $\boldsymbol{c}^{(-1)}$ is initialized with $0$. $\boldsymbol{h}^{(l)}$ is the latent state of the link and node at  $l^{th}$ hop. 
The detailed architecture of the stacked R-GCNs has been shown in \figref{fig:framework} (b), and the formula of GRU can be expressed as
\begin{align}
	&o=\sigma\left(\boldsymbol{W}_{o_{1}} \boldsymbol{h}(t)+\boldsymbol{O}_{o_{1}} \boldsymbol{c}(t-1)+\boldsymbol{b}_{o_{1}}\right), \nonumber \\
	&q=\sigma\left(\boldsymbol{W}_{q_{1}} \boldsymbol{h}(t)+\boldsymbol{O}_{q_{1}} \boldsymbol{c}(t-1)+\boldsymbol{b}_{q_{1}}\right), \nonumber \\
	&\boldsymbol{c}^{\prime}(t)=\tanh \left(\boldsymbol{W}_{h_{1}} \boldsymbol{h}(t)+\boldsymbol{O}_{h_{1}}(q \odot \boldsymbol{c}(t-1))+\boldsymbol{b}_{h_{1}}\right), \nonumber \\
	&\boldsymbol{c}(t)=o \odot \boldsymbol{h}(t-1)+(1-o) \odot \boldsymbol{c}^{\prime}(t)
\end{align}
where $\boldsymbol{W}_{o_{1}}, \boldsymbol{O}_{o_{1}}, \boldsymbol{W}_{q_{1}}, \boldsymbol{O}_{q_{1}}, \boldsymbol{W}_{h_{1}}, \boldsymbol{O}_{h_{1}}$ are the learnable parameters, and $\boldsymbol{b}_{o_{1}},\boldsymbol{b}_{q_{1}},\boldsymbol{b}_{h_{1}}$ are biases. }

\subsection{Multi-task Learning Module}
In this section, we will introduce the productions of MWSL-TTE and the route recovery algorithm together.

\subsubsection{Generating nodes and links travel time}
\textcolor{black}{As we mentioned in \secref{sec:Formulation}, the task of TTE can be formulated as given the real-time $OD$ pairs and the corresponding observation $T$ in $[\mathbf{t},\mathbf{t}+\Delta \mathbf{t}_1)$, we aim to complete the travel time for all links and nodes, and evaluate them using the OD pairs in $[\mathbf{t}+\Delta \mathbf{t}_1,\mathbf{t}+\Delta \mathbf{t})$.  \textcolor{black}{To address the data sparse issue}, this paper formulates it as the problem of tensor completion \cite{li2014missing} by  tensor decomposition technique, 
a popular method for traffic missing value imputation method. Since urban travel time has typical temporal and spatial distribution characteristics, it can frequently be divided into two levels: one is the modeling of road segments or intersections in space, and the other is temporal embedding in time (such as Weather ID and Holiday ID). Based on the above spatio-temporal embedding, we finally employ the 1st order CP decomposition to reconstruct the travel time distribution as}
\begin{align}
	&\mu_e=( \boldsymbol{h}_{e}^{(l+1)}W_{\mu_e}+b_{\mu_e})\otimes \mathcal{T}_i, \ \sigma_e =( \boldsymbol{h}_{e}^{(l+1)}W_{\sigma_e}+b_{\sigma_e})\otimes \mathcal{T}_i, \nonumber \\
	&\mu_v=( \boldsymbol{h}_{v}^{(l+1)}W_{\mu_v}+b_{\mu_v})\otimes \mathcal{T}_i, \ \sigma_v =( \boldsymbol{h}_{v}^{(l+1)}W_{\sigma_v}+b_{\sigma_v})\otimes \mathcal{T}_i, \nonumber
\end{align}
where $\mu_e,\sigma_e \in \mathbb{R}^{|\mathcal{E}|\times 1}$, and $ \mu_v, \sigma_v \in \mathbb{R}^{|\mathcal{V}|\times 1}$ are the mean and variance in Gaussian distribution for link and node respectively. $W_{\mu_e},W_{\sigma_e},W_{\mu_v},W_{\sigma_v} \in \mathbb{R}^{d^{(l)} \times d^{(l+1)}}$,  $b_{\mu_e},b_{\sigma_e} \in \mathbb{R}^{|\mathcal{E}|}$, and $b_{\mu_v},b_{\sigma_v} \in \mathbb{R}^{|\mathcal{V}|}$ are the parameters in the fully connected layer (FC).  $\mathcal{T} \in \mathbb{R}^{  d^{(l+1)} \times I}$ is the embedding tensor of time, and $\mathcal{T}_i \in \mathbb{R}^{  d^{(l+1)} \times 1}$ denotes the embedding vectors in real-time interval.  We discretize the day of time into $I$ time slots (e.g.,  $\Delta t$=15 minutes).  According to \textit{ expected log-likelihood} in \eqnref{equ:obj2}, since the  normal distribution is closed with addition,  the loss function can be derived as 
\begin{align}
	L_{\mu,\sigma}&=-\frac{\left(Q(Z_{1:K})-Q(\mu_{1: K}) \right)^{2}}{2 \sum_{i}(\sigma^{2})^{i}}
	-\frac{1}{2} \log \left(2 \pi  \sum_{i}(\sigma^{2})^{i}\right)\nonumber \\
	&=-\frac{\left(T-Q(\mu_{1: K}) \right)^{2}}{2 \sigma^{2}}
	-\frac{1}{2} \log \left(2 \pi \sigma^{2}\right),
\end{align}
where $Q$ is the aggregation function defined in \defref{def:tte}, and $K$ denotes the number of samples in \textit{bag}. In this paper, \textit{bag} is equal to \textit{route} (\defref{def:route}) between origin-destination.  
\subsubsection{Transition Probability Generative Layer}  
\textcolor{black}{
We thereafter introduce the detailed structure of the transition probability generative layer  to generate the link transition probability by using edges features $\boldsymbol{h}^{(l+1)}$ (\eqnref{equ:stack_gcn}). Technically, for the last two layers, we use the multi-layer perceptron (MLP) to produce the weights of links 
\textcolor{black}{
\begin{align*}
	a_{i \rightarrow j}=\operatorname{MLP}\left( \boldsymbol{h}_{i}^{(l+1)} \ || \ \boldsymbol{h}_{j}^{(l+1)}\right),
\end{align*}}
where $||$ is the operator of features-wise concatenation,  and then apply the softmax layer over outgoing links
\begin{align}
	p(e_j \mid e_{i})=\frac{\exp(a_{i \rightarrow j})}{\sum_{v_{n} \in \mathcal{V}_{i}} \exp(a_{i \rightarrow n})}\label{equ:transition}
\end{align}
After that, the transition probability in \eqnref{equ:transition}  will be employed in the Route Search Layer. For simplicity, we use the transition matrix $A \in \mathcal{R}^{|\mathcal{V}|\times |\mathcal{V}|}$ to represent all links' possibilities; for example, we have $A[i,j]= P(e_j \mid e_{i})$.}

\subsubsection{Route Search Layer}\label{sec:route_search}
\textcolor{black}{
As aforementioned, the shortest route algorithms omit the condition of the road in practice. To solve this problem, we intend to construct the transition probability between road segments and combine the transition probability to infer the route. However, considering the complex highway graph, there are a tremendous number of routes for any OD pair.  It would be reasonable to  prune the routes through some thresholds. Therefore, in this paper, we prune the route from two aspects: 1) The lengths of the simple route $r$ from origin $o$ to destination $d$ should satisfy $r.length < r_{short} + \delta_{lens}$, where $r_{short}$ is the shortest simple route and $\delta_{lens}$ is the distance threshold. 2) the co-occurrence probability of a simple route $P(r|OD; A)$ should also meet the criteria $P(r|OD; A)>\delta_{probs}$, where $\delta_{probs}$ is the probability threshold. Next, we will introduce the definition of probability $P(r|OD; A)$. 
 According to Assumption 2 and \eqnref{equ:transition}, the co-occurrence route probability 
\begin{align}
	p(r)=P(e_1,e_2,\cdots e_{K})= p(e_0) \prod_{i=1}^{K} p(e_i \mid e_{i-1}),\label{equ:route_probs}
\end{align}
where $e_0$ is the origin location, and we have $p(e_0)=1$.}

\textcolor{black}{After the strategy of pruning has been introduced, the candidate routes can be obtained by  the Depth First Search (DFS) algorithm. Specifically, we  generate the candidate route set $\Omega_{OD}$ via posterior probability $p(r|OD; A)$ and choose the route with maximum  probability as the optimal solution. Formally, given an OD pair and observation $T$, the optimal route $r^*$ can be written as
\begin{align}\label{eqn:route}
r^*&=\mathop{\arg\!\max_{r_i \in \Omega_{OD}}} p(r_i \mid OD; A;  T; Z) \nonumber \\
&=\arg \min_{r} |  T - \sum_{e_i \in r_j}\mu_e^{(i)}- \sum_{v_i \in r_j}\mu_v^{(i)} |, \ \forall r_j \in \Omega_{OD}.
\end{align}
Eq. \ref{eqn:route} selects the most suitable route $r^*$ regarding current travel time $\mu$. }


\subsubsection{Model Training}
\textcolor{black}{
Next,  we  will introduce  the optimizing  procedure of MWSL-TTE. As we discussed in \secref{sec:route_search}, the top $m$ maximum routes have been obtained, and we chose the most 
satisfied one by \eqnref{eqn:route}.  However, such a choice may fall into a local solution, and other candidate routes might never be picked. To address this problem, we here introduce the $\epsilon$-greedy algorithm, which means that  the route satisfied \eqnref{eqn:route} will be chosen in $1 - \epsilon$ probability. Otherwise, randomly select the routes with top $m$  maximum probability in $\epsilon$ probability.} Moreover, we wish that the transition probability could help us infer the most possible route based on the ground truth (observation travel time). So, we here adopt the Kullback-Leibler (KL) divergence to measure the coherence, which can be written as  
\begin{align}
	D_{\mathrm{KL}}(\mathcal{P} \| \mathcal{Q})=-\sum_{i} \mathcal{P}(i) \ln \frac{\mathcal{Q}(i)}{\mathcal{P}(i)},
\end{align}
where $\mathcal{P}$ is the probability distribution of each route in the candidate set, and $\mathcal{Q}$ is the inverse estimation loss distribution between $Q(\mu)$ and ground truth. In other words, the optimization direction is towards both higher route probability and more accurate travel time estimation. For the route picked up through \eqnref{eqn:route}, the Negative Log
Likelihood (NLL) loss has been employed to minimize the negative log
likelihood function, which can be defined as
\textcolor{black}{
\begin{align}
	L_{tp}=-\sum_{i=2}^{|t_r|}\log(p(e_i \mid e_{i-1}, \theta)),
\end{align}
where $\theta$ is the model's trainable parameters to represent the posterior probability. }
By fusing all objective functions together, 
our model is trained to minimize the weighted combination of three loss terms
\begin{align}\label{equ:objs}
	\mathcal{L}=\alpha L_{\mu,\sigma}+ \beta L_{tp} +(1-\alpha-\beta) D_{\mathrm{KL}}(\mathcal{P} \| \mathcal{Q})
\end{align}
where $\alpha$ and $\beta$ are the const parameter to balance three loss terms $L_{\mu,\sigma},L_{tp}$ and $D_{\mathrm{KL}}(\mathcal{P} \| \mathcal{Q})$. 
The training pseudocode of  MWSL-TTE has been depicted in Algorithm \ref{alg:training}.

\begin{algorithm}
 \caption{Training Procedure of MWSL-TTE}
  \label{alg:training}
\LinesNumbered
\KwIn{OD datasets $\mathcal{D}$, node-wise graph $\mathcal{G}_v$, link-wise graph $\mathcal{G}_e$, and number of candidates routes $N$}
\KwOut{ OD TTE estimation function $f$}

\While{not convergence}{
	
	$(\mu,\sigma,A) = f(W; \mathcal{G}_v; \mathcal{G}_e)$ \\ 
	\For{$OD \in \mathcal{D}$}{
		\begin{itemize}[leftmargin=*]
			\item[1.] Generating Top $N$ candidates route set $\Omega_{OD}$ \\ 
			\item[2.] Select the route $r_i$ from $\Omega_{OD}$ through $\epsilon$-greedy
			\item[3.] Calculate  the  loss  by  \eqnref{equ:objs} and update the \\ parameters through back-propagation.
		\end{itemize}

}
}

\textbf{return}  $f$
\end{algorithm}
\section{Experiments}\label{sec:exps}
In this section, various experiments will be conducted based on a wide range of public real-world taxi dataset in Xi’an, and Chengdu to evaluate the superiority of MWSL-TTE in TTE and  route recovery aspects.
\subsection{Datasets}
\noindent\textbf{Road Networks.} We use two road networks: Chengdu Road Network and Xi'an Road Network. \textcolor{black}{Both of them are extracted from OpenStreetMap \cite{haklay2008openstreetmap}, and include nine road types (trunk, trunk link, freeway link, primary, primary link, secondary, secondary link, tertiary, tertiary link). Here, Chengdu road network contains 8221 edges and 5182 nodes, which ranges from 30.63\degree{} to 30.69\degree{} in latitude and 104\degree{} to 104.07\degree{} in longitude.  And Xi'an road network contains 4780 edges and 3782 nodes, ranging from 34.20\degree{} to 34.29\degree{} in latitude and 108.90\degree{} to 108.99\degree{} in longitude.}

\noindent\textbf{Taxi OD Orders.} We use two public taxi trajectory datasets come from the Didi Express platform to generate the OD orders. Each generated order corresponds to a trip record that consists of the time-stamps and locations of an OD. Here, we implement Xi'an dataset that is from 10/10/2016 - 10/22/2016, and Chengdu Dataset is from 08/18/2014 - 08/24/2014 (a whole week from Monday to Sunday). The GPS points of both two datasets have been tied to the road and the interval of sample trajectory points is 2-4s, ensuring that the vehicle trajectory can correspond to the actual road information. Especially, we generate the ground truth route of the original vehicle trajectories for the route recovery task via a map-matching tool FMM \cite{yang2018fast}.

 \begin{table*}[tb]
	\centering
	
	\caption{Performance of MWSL and its variants for OD travel time estimation, compared with other baseline methods. } 
	\vskip -0.1in
	\setlength{\tabcolsep}{2mm}{	
		\begin{tabular}{l  c c c | c c c }
			\toprule \multirow{2}{*}{Models}   & \multicolumn{3}{c}{Xi'an} & \multicolumn{3}{c}{Chengdu} \\
			\cline{2-4} \cline{5-7} 
			& {\small RMSE (sec)} & {\small MAE (sec) } & {\small MAPE} & {\small RMSE (sec)} & {\small MAE (sec) } & {\small MAPE} \\
			\midrule
			{TEMP}& 398.95&277.56&34.24\%&446.98&327.99&32.00\%\\
			{GBDT} & 365.72&250.63&31.27\%&435.88&303.83&30.35\%\\
			{STNN} &353.06 &241.19&30.43\%&425.53&293.10&28.25\%\\
			{MURAT}&538.23&512.65&127.87\%&519.20&503.36& 118.62\%\\
			{DCRNN}&282.54&191.41&24.94\% &392.45&263.91&25.95\%\\
			{ConSTGAT}&283.89&195.31&25.32\%&403.31&280.90&28.16\%\\
            {ASTGNN}&259.46&179.08&23.86\%&362.48&244.03&23.52\%\\
			\hline
			{N-Node}& 253.62&173.71&23.04\% &367.04&236.18&23.69\% \\
			{N-GRU}& 259.52&181.37&24.25\%&371.34&240.45&24.03\% \\
			{N-R-GCN}&263.46 &184.73&24.60\%&364.58&234.37&23.46\% \\
			{N-PathUpdate}&257.62&178.25&23.59\%&358.09&229.56&23.15\% \\
			{MWSL-TTE } &$\textbf{238.86}$&$\textbf{162.37}$&$\textbf{21.33\%}$&$\textbf{341.02}$&$\textbf{215.03}$&$\textbf{22.27\%}$\\
			\bottomrule
	\end{tabular}}

	\label{table:result}
\end{table*}

\subsection{Baseline Methods and Metrics} We first compare our models with six baseline methods for the task of OD travel time estimation:
\begin{itemize}[leftmargin=*]

\item \textbf{TEMP}: Temporally weighted neighbors \cite{wang2019simple} is a nearest-neighbor-based approach that estimates the OD travel time by averaging the travel time of all historical trajectories falling in the same time slot with a similar origin
and destination.

\item \textbf{GBDT}: Gradient boosting decision tree \cite{friedman2001greedy} is used for the regression of OD travel time estimation.

\item \textbf{STNN}: Spatio-temporal deep neural network\cite{jindal2017unified} is a deep neural network-based approach that first predicts the travel distance given an OD pair, and then combines this prediction with the departure time to estimate the
travel time. 

\item \textbf{MURAT}: Multi-task representation learning \cite{li2018multi} is a deep neural network-based approach that jointly predicts the travel distance and the travel time for taxi orders by learning representations of road segments and the origin-destination information.

\item \textbf{DCRNN}: It exploits GCN to capture spatial dependency, and then uses recurrent neural networks to model temporal dependency \cite{li2017diffusion}. We implemented this model based on OD estimation prediction of road network. \textcolor{black}{The hidden vector size of GCN and GRU are set as 20 and 128, respectively.}

\item \textbf{ConSTGAT}: This model adopts a graph attention mechanism to explore the joint relations of spatio-temporal information \cite{fang2020constgat}. \textcolor{black}{The parameter setting is basically same with the original model. In the integration module, we also use two-layer MLP.}

\item \textbf{ASTGNN}: This model consider multiple factors in traffic forecasting, such as, periodicity, spatial
heterogeneity by leveraging a trend-aware multi-head attention mechanism \cite{guo2021learning}. \textcolor{black}{The number of layers for both encoder and decoder is set to 3. And the kernel size of convolution is set to 5.}

\end{itemize}

Moreover, for the task of OD route recovery, we compare with two representative baselines of route recovery from sparse trajectories. Both of them are based on inverse reinforcement learning to capture the spatial transition probabilities, and the difference between these two models is that the temporal components:

\begin{itemize}[leftmargin=*]
\item \textbf{STRS}: Spatio-temporal-based route recovery system \cite{wu2016probabilistic} seeks to recover the route from sparse trajectories. The temporal components of STRS comprise a matrix factorization-based method.

\item \textbf{DeepGTT-STRS}: Li et al. \cite{li2019learning} proposes a deep generative travel time estimation model named DeepGTT that replaces the temporal component of STRS. 

\end{itemize}

\noindent\textbf{Evaluation Metrics.} For the OD travel time estimation of our MWSL, we evaluate the performance with RMSE (Root Mean Square Error),
MAE (Mean Absolute Error), and MAPE (Mean Absolute Percentage Error). Then we adopt the accuracy of route recovery as the main performance metric for the route recovery task. It is defined as the ratio of the length of a correctly inferred route to the length of the ground truth route $R_G$ or the inferred route $R_I$ whichever is longer, i.e., $accuracy = \frac{(R_{G}\cap R_{I}).len} {max\{R_{G}.len,R_{I}.len\}}$. 

\subsection{Experimental Settings}

 The experiments are implemented with PyTorch 1.6.0 and Python 3.6, and trained with a RTX2080 GPU. The platform ran on Ubuntu 16.04 OS. We trained the models using Adam optimizer with an initial learning rate of 0.001 on both Chengdu and Xi'an datasets, and early stopping is used on the validation dataset. Especially, we run each experiment for three times.
 
 \textcolor{black}{The main hyper-parameter settings of our proposed method are described as follows:}

\begin{itemize}
	\item[$\bullet$] \textcolor{black}{In the generation of a candidate route set $\Omega_{a,b}$, $m$ candidate routes are selected between the OD pairs. Here, $m=6$ is used for Xi'an, and $m=8$ for Chengdu, respectively. Both two hyper-parameters can ensure that over 90 percents of ground truth route can be acquired from $\Omega_{OD}$.}
	
	\item[$\bullet$] \textcolor{black}{The number of stacked R-GCNs is set to 3.}

	\item[$\bullet$] \textcolor{black}{In the road attributes embedding layer, the embedding sizes of link feature representation (road ID, road types, number of lanes and one way or not) are set to 128, 8, 4 and 2, respectively. And the embedding size of node feature representation (node ID, node type and node street count) are set to 96, 2 and 2, respectively.}
	\item[$\bullet$] \textcolor{black}{In the temporal embedding component, we embed Weather ID and Holiday ID in $R^8$ and $R^4$, respectively.}
\end{itemize}

\subsection{Experimental Results}

 We compare our MWSL-TTE with other baseline methods under two datasets.

\begin{table}[tb]
	\centering
	\caption{\textcolor{black}{Inference time cost comparison for GCN-based travel time estimation models. The time unit is second.}} 
	\vskip -0.1in
	\setlength{\tabcolsep}{1.5mm}{
	\begin{tabular}{l  c  | c  }
		\toprule \multirow{2}{*}{Models}    & {Xi'an} & {Chengdu}  \\
		\cline{2-3}
	& {\small Time (sec) } &  {\small Time (sec)} \\
		\midrule
		{DCRNN}& 0.15&0.16  \\
		{ConSTGAT}& 0.41&0.45 \\
	    {ASTGNN}& 0.34&0.37 \\
		{MWSL-TTE }&0.23&0.25\\


		\bottomrule
	\end{tabular}}
	
	\label{table:time}
\end{table}

\begin{table}[tb]
	\centering
	\caption{Performance of MWSL and STRS-based baseline methods for route recovery. The column T.time refers to the training time of the model and its unit is hour. } 
	\vskip -0.1in
	\setlength{\tabcolsep}{1.5mm}{
	\begin{tabular}{l  c c | c c  }
		\toprule \multirow{2}{*}{Models}   & \multicolumn{2}{c}{Xi'an} & \multicolumn{2}{c}{Chengdu}  \\
		\cline{2-3}  \cline{4-5} 
		& {\small Acc} & {\small T. time} & {\small Acc} & {\small T. time} \\
		\midrule
		{STRS}& 82.71\%&3.18&71.64\%&4.74  \\
		{DeepGTT-STRS}& 79.39\%&3.37&68.72\%&5.02 \\
		
		{MWSL-TTE }&\textbf{86.25}\%&\textbf{0.52}&\textbf{77.03}\%&\textbf{0.69}\\


		\bottomrule
	\end{tabular}}
	
	\label{table:path}
\end{table}

\begin{table*}[tb]
	\centering
	\caption{Performance of MWSL under different combinations of $\alpha$ and $\beta$ based on Xi'an and Chengdu dataset for both OD travel time estimation and route recovery.} 
	\vskip -0.1in
	\setlength{\tabcolsep}{1.5mm}{	
	
	\begin{tabular}{l  c c c |  c  }
	    	\toprule {Datasets}  & \multicolumn{4}{c}{Xi'an/Chengdu} \\
		\toprule \multirow{2}{*}{Parameters}   & \multicolumn{3}{c}{TTE} & \multicolumn{1}{c}{Route recovery }  \\
		\cline{2-4}  \cline{4-5}
		& {\small RMSE (sec)} & {\small MAE (sec) } & {\small MAPE (\%)} & {\small Acc (\%)}  \\
		\midrule
		{($\alpha$=1,$\ \beta$=0)}&240.62/354.36&161.12/233.27&21.62/23.17& \verb|\| \\
		{($\alpha$=0.8,$\ \beta$=0.2)}&243.63/347.48 &164.15/221.19&20.86/22.64&85.94/73.36\\
		{($\alpha$=0.8,$\ \beta$=0.1)}&238.86/341.02 &162.37/215.03&21.33/22.27&86.25/77.03\\
		{($\alpha$=0.6,$\ \beta$=0.2)}&248.41/352.29 &167.56/232.34&22.06/22.83&86.92/78.14\\
		{($\alpha$=0.6,$\ \beta$=0.3)}&247.46/358.75&166.55/239.49&22.12/23.39&84.17/73.22 \\
		{($\alpha$=0.6,$\ \beta$=0.1)}&249.00/365.27&166.59/237.92&21.84/23.78&80.83/67.50 \\
		\bottomrule
	\end{tabular}	}

	\label{table:hyper}
\end{table*}

\begin{figure}[t]
	\centering
	\setlength{\belowcaptionskip}{-0.3cm}
	\includegraphics[width=1\linewidth]{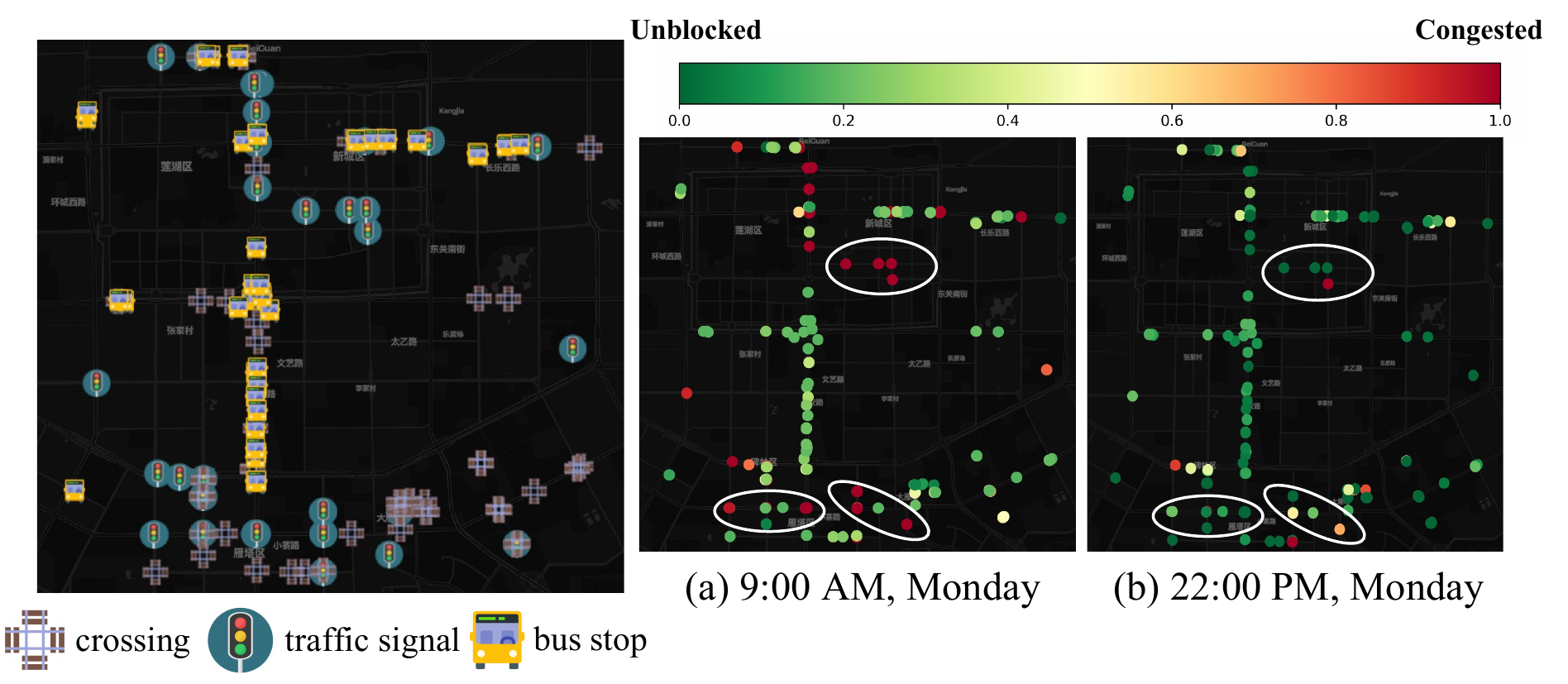}
	\caption{The time-consuming ratio for some nodes in the Xi'an road network. Here, we select three types of nodes including "crossing", "traffic signal" and "bus stop".}
	\label{fig:node}
\end{figure}

\begin{figure}[t]
	\centering
	 \setlength{\belowcaptionskip}{-0.1cm}
	\includegraphics[width=0.9\linewidth]{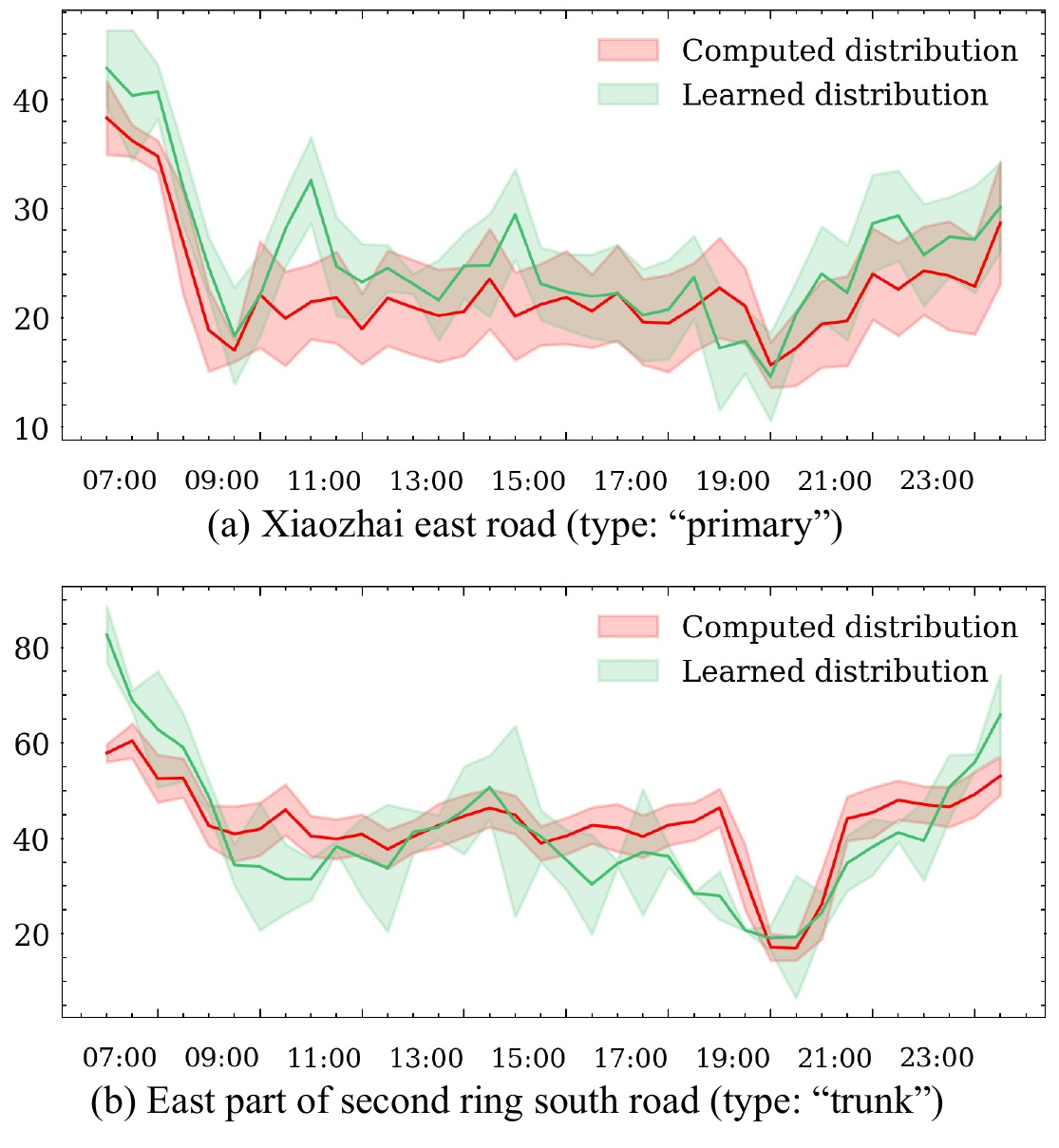}
	\caption{Learned speed distributions of two sample links by MWSL, compared with the speed distributions computed by the original taxi trajectories.}
	\label{fig:link}
\end{figure}
\noindent\textbf{Performance on Travel Time Estimation.} \tableref{table:result} shows the overall performance of estimating the travel times of Taxi OD orders. From the performance comparison, we find that our MWSL-TTE achieves the best performance than other methods in terms of all three metrics. The better prediction results can be explained in two aspects. First, our model implements an effective graphical structure to capture the prior information of road network. \textcolor{black}{Although all DCRNN, ConSTGAT and ASTGNN model the link-wise adjacency of road segments, the node-wise features especially traffic signal junctions are ignored. Thus, these three models without considering the node-wise adjacency cannot achieve higher accuracy.} Second, given an OD pair, our weak supervision-based method can study the travel time distributions of the links and nodes. Compared with Taxi OD estimation baselines such as STNN and MURAT, in-process information of OD pairs improves our estimation results.

\noindent\textbf{Ablation Study.} In \tableref{table:result}, except for the comparison experiment with the six baseline methods, we also conduct the ablation study by replacing our MWSL-TTE with four variations, namely N-Node, N-GRU, N-R-GCN and N-PathUpdate, to evaluate the
effectiveness of different modules in MWSL-TTE (see \figref{fig:framework}). In N-Node, we remove the node-wise estimation. In N-GRU, we remove the stacked GRU and only use the same number of layers of R-GCN. In N-R-GCN, we remove the R-GCN and replace it with the same number of layers of normal GCN \cite{kipf2016semi}. In N-PathUpdate, we only implement the initial path as the in-process trajectories of OD orders and do not update the path based on the learned travel time and transition probability. The result comparison, it shows that R-GCN and node-wise estimation are the most critical parts. Regardless of the node-wise aspect, the performance becomes worse, and it proves the importance of modeling the complex adjacency for both the links and nodes. Furthermore, stacked R-GCNs with GRU integrate the multiscale information to capture both global and local features, and path update is also important in improving the travel time estimation. To sum up, the key designs of MWSL-TTE are effective. 
\begin{figure*}[t]
	\centering
	\includegraphics[width=1\linewidth]{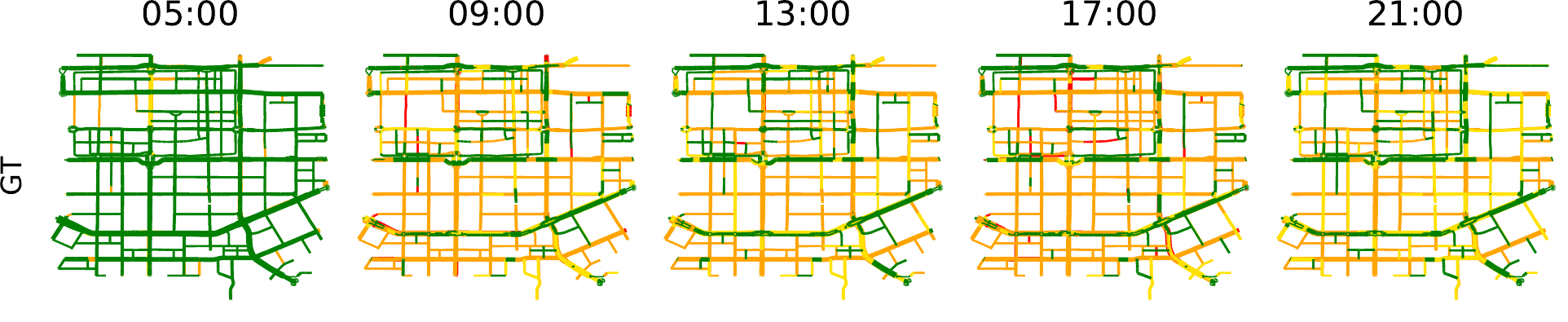}
	
	\includegraphics[width=1\linewidth]{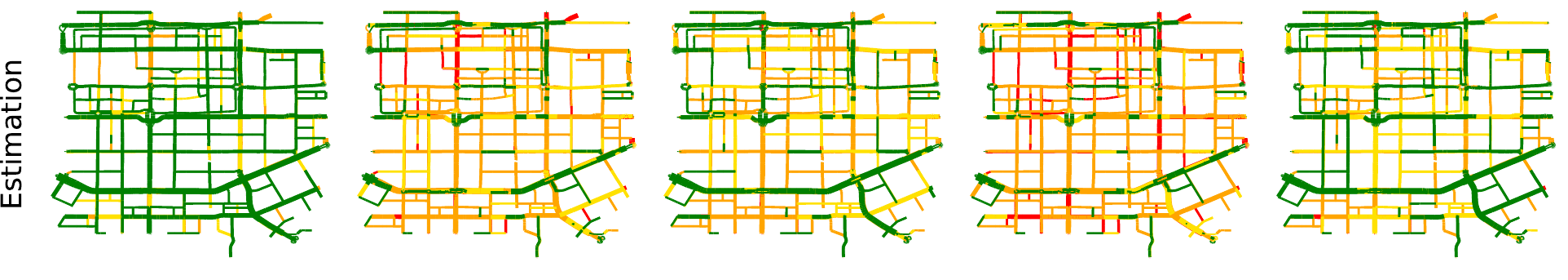}
	\caption{Comparison between ground truth traffic state and estimated traffic state from MWSL-TTE. Here we use four kinds of colors to represent the different road states, which can be defined as 1) red - very congested, 2) orange - congested, 3) yellow - slow, and 4) green - unblocked.}
		\label{fig:road_network}
\end{figure*}
\begin{figure}[!b]
    \centering
    \subfigure[Xi'an.]{
        \label{figure:81a}
        \includegraphics[height=0.23\textwidth]{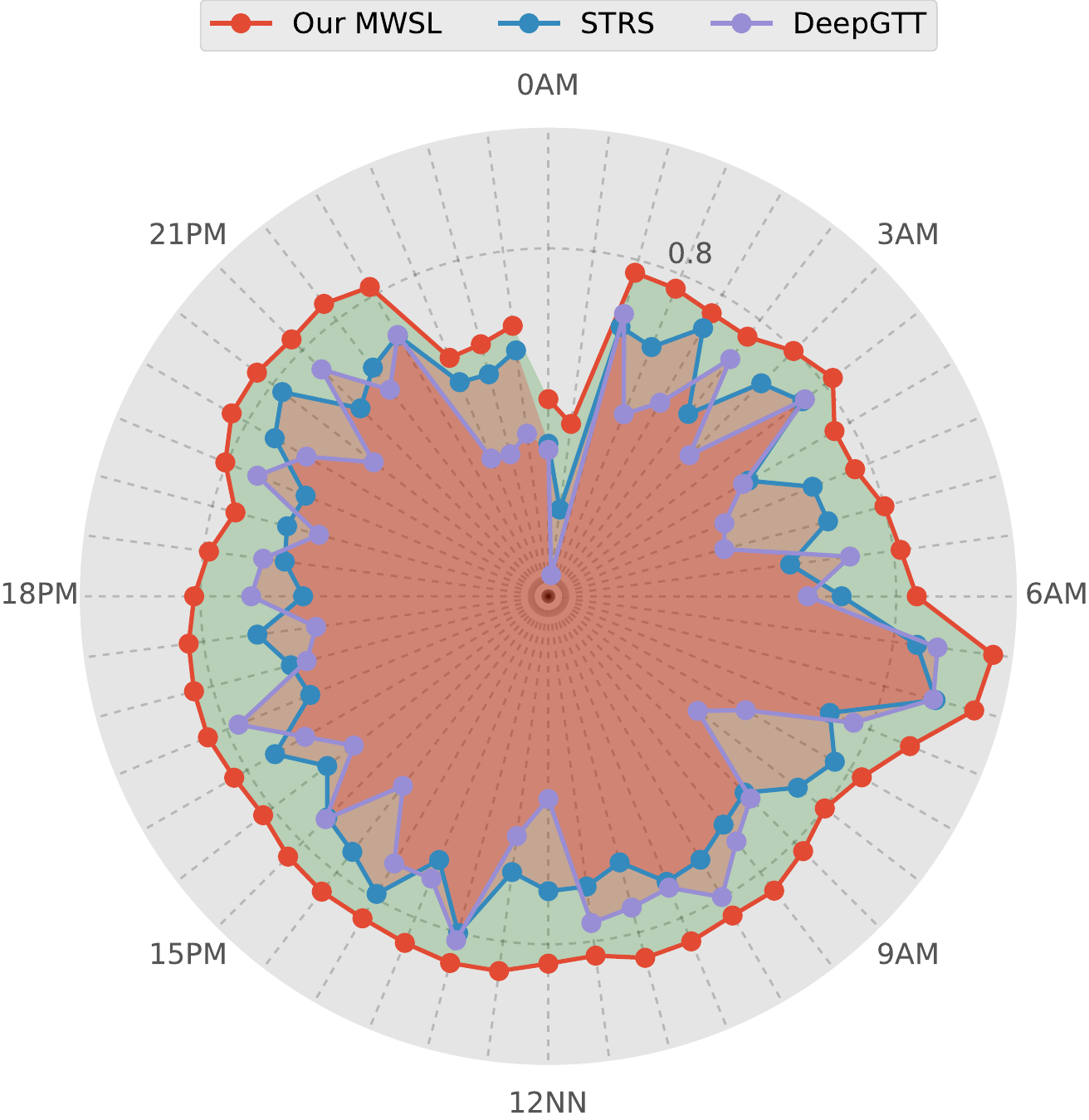}
    }
    \subfigure[Chengdu.]{
        \label{figure:82b}
        \includegraphics[height=0.23
        \textwidth]{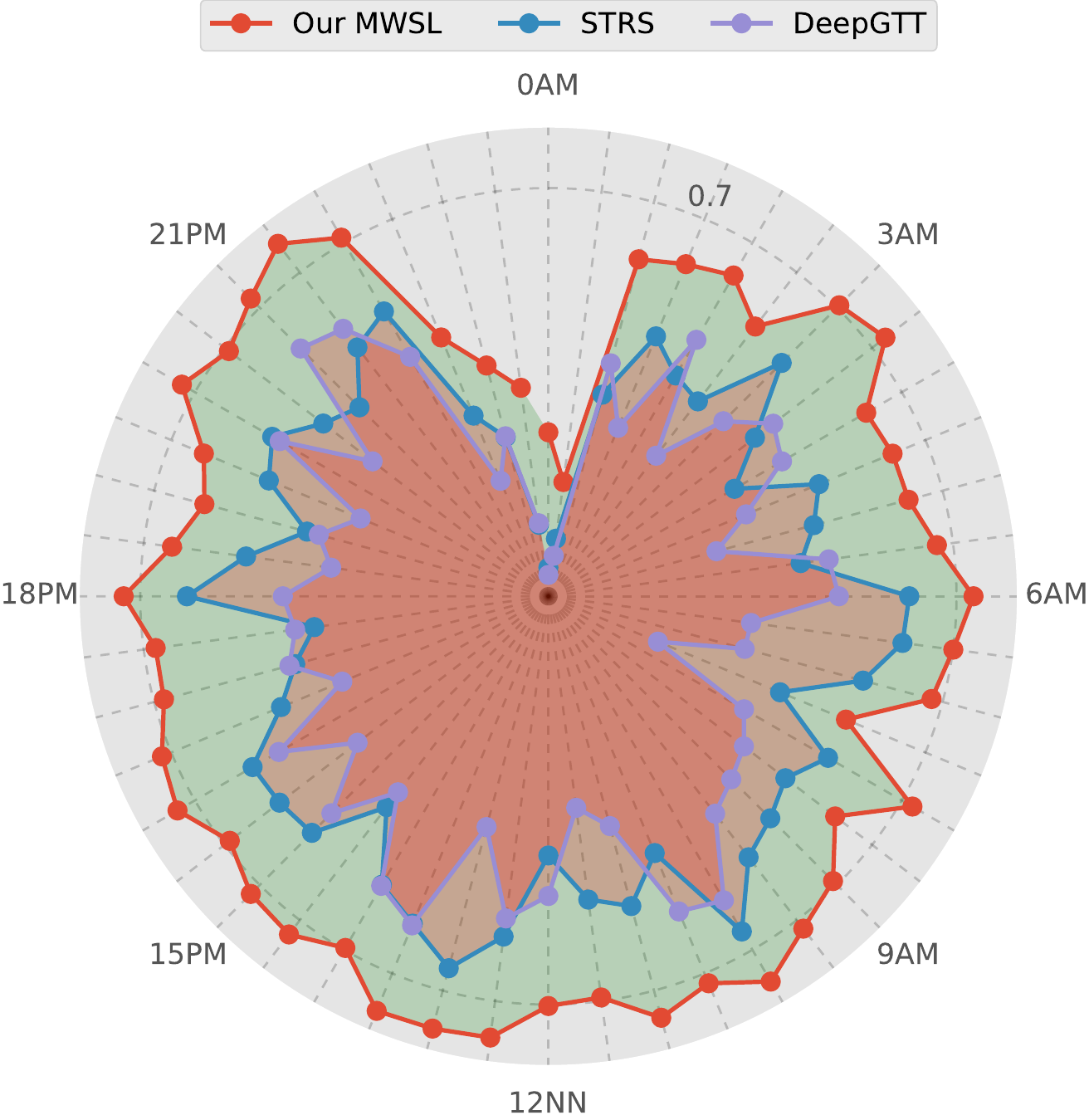}
    }
    \caption{The 24-h divergence between generated road conditions and ground truth under datasets of both Xi'an and Chengdu, compared with the temporal components of two baseline methods.}
    \label{figure:divergence}
\end{figure}

\textcolor{black}{\noindent\textbf{Time Cost Analysis.} Due to the online travel time estimation of our proposed MWSL-TTE, we further compare the inference time with main GNN-based baselines. As \tableref{table:result} shows, GNN-based travel time estimation models have significantly better performance than common deep learning-based methods, with a batch size 32 using a RTX2080 GPU card. \tableref{table:time} provides the inference time cost comparison for main GNN-based models. Although our proposed model is slower than DCRNN, the inference time of our model can acquire a faster inference time comapred with other relatively complex models. And this inference time result also proves that our model can process an online travel time estimation. }

\noindent\textbf{Performance on Route Recovery.} \tableref{table:path} shows the performance of the route recovery task by given OD pairs. We can find that our path recovery of MWSL has better recovery accuracy and shorter model training time. STRS-based methods are very time-consuming due to the long iterations of inverse reinforcement learning to acquire the transition probability among road segments. Observe that the accuracy of Deep-STRS is always worse than that of STRS. The reason is that a larger sampling time interval between OD pairs leads to a more inaccurate grid-based traffic tensor which is the input of DeepGTT to model the traffic representation. In particular, a more complex Chengdu road network leads that the recovery accuracy of three methods dropping as expected.

\noindent\textbf{Hyper-parameter Analysis.} To further show the effectiveness of multi-task components of our model, we conduct experiments under different combinations of parameters $\alpha$ and $\beta$ based on both of the two datasets. As observed in \tableref{table:hyper}, on one hand, we find that in terms of the TTE task, the overall TTE performance improves as $\alpha$ changes from 0.6 to 0.8 under the datasets of two cities. However, $\alpha=1$ doesn't achieve the best TTE performance. This is because that the $\beta$ that controls the loss terms of route recovery also has some impacts on TTE prediction. More accurate path updates can bring the improvement of TTE prediction. On the other hand, the route recovery performance achieves the best accuracy when $\alpha=0.8$ and $\beta=0.1$. This indicates that the $D_{\mathrm{KL}}(\mathcal{P} \| \mathcal{Q})$ also plays its part in obtaining a higher accuracy for route recovery. Furthermore, we compare the route recovery performance when $\alpha=0.6$. The optimal hyper-parameter is the combination of $L_{tp}$ and $D_{\mathrm{KL}}(\mathcal{P} \| \mathcal{Q})$ as well, but excessive loss weight of $D_{\mathrm{KL}}(\mathcal{P} \| \mathcal{Q})$ would cause the poor prediction performance. To sum up, we conclude that the experimental results demonstrated the superiority and generality of the multi-task components of our proposed MWSL.

\section{Case Study}\label{sec:case_study}
Our MWSl-TTE not only can conduct path travel time estimation and route recovery for OD GPS trips, but also learn the travel time distributions of the links and nodes based on weakly supervised learning. Thus, in addition to the quantified evaluations described in Section 6.4, we also conduct a real-world case study in Xi'an, which visualizes the learned distributions of the links and nodes from the road network. Especially, we conducted the comparison with road condition computed by original taxi trajectories.

\subsection{Learned Distributions for Nodes and Links} 

On one hand, we first provide the estimation results for different types of nodes, which are depicted in \figref{fig:node}. We select three types of nodes in a road network and use the time-consuming ratio to represent congestion status. The time-consuming ratio is calculated by dividing the corresponding mean value of learned travel time distributions by the maximum mean value among all nodes (Noted that we filter out top 1\% largest node travel time). From \figref{fig:node}, we can find that nodes with the "traffic signal" type are more time-consuming than the other two types of nodes, and most nodes in the morning peak are easy to become congested. These indicate that our node travel time estimation is reasonable in both spatial and temporal aspects.

On the other hand, we transform the travel time distribution into speed distributions in terms of links, and this transformation process is based on \cite{li2015inferring}. The reason for this transformation is that most users drive cars with a normal speed range (e.g., 10 kmph to 120 kmph), and thus we can easily analyze the rationality of learned speed distribution compared with link travel time. Two links' speed distributions were generated by the proposed MWSL, as is shown in \figref{fig:link}, and we compared them with the speed distribution that is computed by the original taxi trajectories. To test the generalization of our model, we select two types of links, where the Xiaozhai east road is a busy link (type: "primary"), and we can find that the morning and evening peaks are obvious in both learned speed distribution and computed distribution. Another link is a highway link, and the commuting pattern only appears in the evening for both distributions. Based on the above analysis, it is concluded that the learned distributions of links can effectively represent different functional types of links. Furthermore, the mean values $\mu$ of learned speed distributions are closer to the ground truth.


\subsection{A Demo of the Generated Road Conditions}
We generate the road conditions by our MWSL based on taxi OD trips, and we compare with the ground truth computed by the original taxi trajectories. Especially, we mark it with the unblocked state for the road segments without taxi trajectories. Since the speed limit for each road is varied, which is primarily defined by road type or road length, we use four kinds of colors to represent the different road states (very congested, congested, slow and unblocked). We divide the  limiting-velocity for each road type equally, for example, the rate-limiting of primary road is $60kph$, so the interval between very congested is $[0, 15)$, congested is $[15, 30)$ ,slow is $[30, 45)$ ,unblocked is $[45, 60)$. The compared result is shown in \figref{fig:road_network}. From the comparison of the generated road condition and ground truth at several time slots, we can acquire the following insights: 1) similar traffic state. The road condition generated by our model is similar to the ground truth. Most road segments have the same road states, and those road segments with different road states frequently have a consistent tendency; 2) rational adjacency correlation. We can find that the road segments with neighbor segments often have the similar road state for the generated road condition map. This indicates that our model can learn adjacency correlation of road network.

To better illustrate the model performance for generated road conditions, we also conduct quantitative analysis under Xi'an and Chengdu datasets. As is shown in  \figref{figure:divergence}, we computes the 24-h divergence between generated road conditions and ground truth. The prediction accuracy is relatively worse among these three methods from 22:00 PM to 2AM. This is because that a very small number of OD pairs in these time slots can not provide efficient model training for better prediction performance. However, the plots show that the generated conditions achieve the accuracy of around 80\% and 70\% at ordinary time slots under the datasets of both Xi'an and Chengdu, respectively. Compared with ground truth, our weak supervision learning method can provide believable road conditions only relying on the OD pairs.

\section{CONCLUSION}\label{sec:concluds}

For the first time, we consider the OD travel time estimation as an inexact supervision problem and propose a multi-task framework to infer the optimal route based on transition probability and learn the travel time distribution for each road segment and intersection through \textit{expect MLE framework} \cite{zhang2020learning}. The stacked R-GCN architecture has been employed to learn the complex relations of the road network, and we generate the travel time distribution for both road segments and intersections by 1st-order CP decomposition. Finally, we produce the transition probability between road segments by multi-layer perception. Moreover, an iterative update strategy has been proposed to update the transition probability and candidate paths during the training process. We evaluate our model on two real-world public datasets and verify the effectiveness of our proposed algorithm.

Future work can be concluded in four parts. Firstly, more potential superior distribution can be developed under the assumptions of weakly supervised learning, since in this paper, only log-normal distribution has been employed. Secondly, more advanced route search algorithms could be designed based on, for example, transition probability, travel time, or route probability. Thirdly, more urban scenarios, such as buses, subways, and people, can be tried to extend the applications of weakly supervised learning or travel time distribution. 
Lastly, a federated learning-based method can be designed, since the OD types of data save abundant storage costs on the client's mobile phone.

\section{Acknowledgment}

We are grateful to anonymous reviewers for their helpful comments.  This work was partially  supported by the grants of National Key Research and Development Program of China (No. 2018AAA0101100), National Key Research and Development Project (2021YFB1714400) of China and  Guangdong Provincial Key Laboratory (2020B121201001).
%
\bibliographystyle{IEEEtran}
\bibliography{reference}
\begin{IEEEbiography}[{\includegraphics[width=1in,height=1.25in,clip,keepaspectratio]{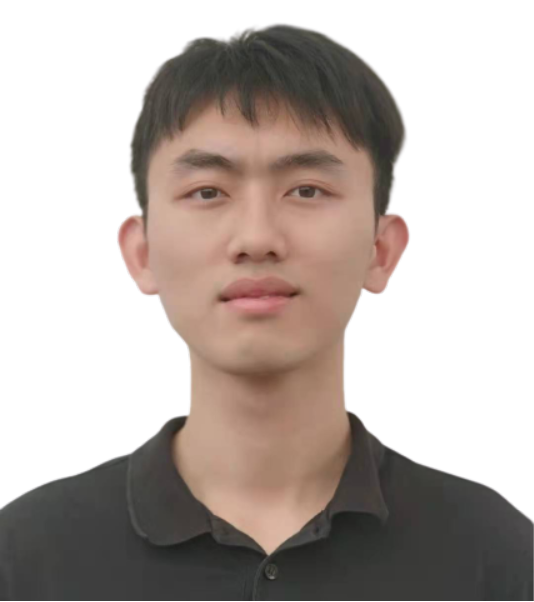}}]{Hongjun Wang} is working toward the M.S. degree
	in  computer science and technology from Southern University of Science and Technology, China. He received the B.E. degree from the Nanjing University of Posts and Telecommunications, China, in 2019. His research interests
	are broadly in machine learning, with urban computing, explainable AI, data mining, data visualization.
\end{IEEEbiography}
\vspace{1ex}
\begin{IEEEbiography}[{\includegraphics[width=1in,height=1.25in,clip,keepaspectratio]{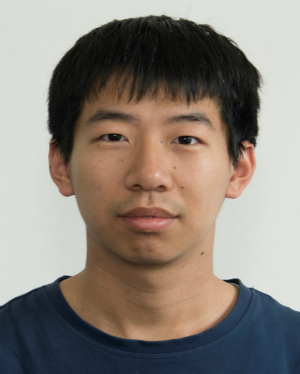}}]{Zhiwen Zhang}  received the B.E. and M.S. degree in Artificial Intelligence from Nankai University, China, in 2016 and 2019 respectively. From 2019, he is currently pursuing a Ph.D. degree at the Department of Socio-Cultural Environmental Studies, The University of Tokyo. His current research interests include urban computing and data visualization.
\end{IEEEbiography}
\vspace{1ex}
\begin{IEEEbiography}[{\includegraphics[width=1in,height=1.25in,clip,keepaspectratio]{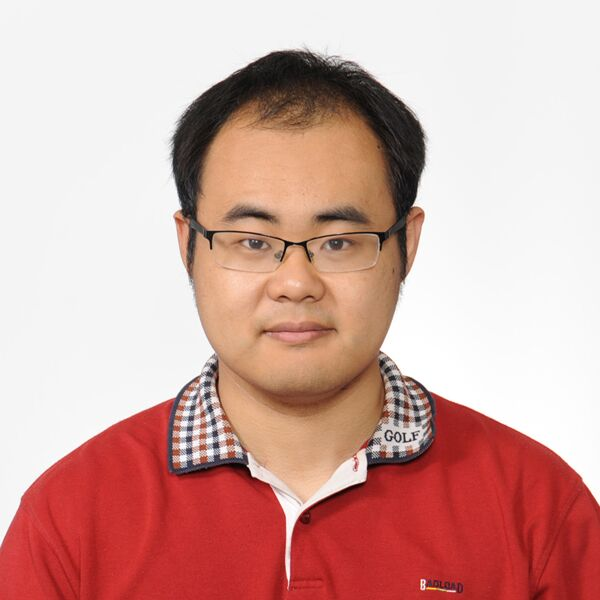}}]{Zipei Fan}  received his B.S. degree in Computer Science from Beihang University, China, in 2012, both M.S. and a Ph.D. degree in Civil Engineering from The University of Tokyo, Japan, in 2014 and 2017 respectively. He became Project Researcher and Project Assistant Professor in 2017 and 2019, and he has promoted to Project Lecturer at the Center for Spatial Information Science, the University of Tokyo in 2020. His research interests include ubiquitous computing, machine learning, Spatio-temporal data mining, and heterogeneous data fusion.
\end{IEEEbiography}
\vspace{1ex}
\begin{IEEEbiography}[{\includegraphics[width=1in,height=1.25in,clip,keepaspectratio]{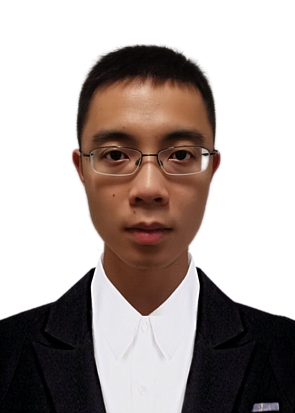}}]{Jiyuan Chen} is working towards his B.S. degree in Computer Science and Technology from Southern University of Science and Technology, China. His major research fields include artificial intelligence, deep learning, urban computing and data mining. 
\end{IEEEbiography}
\vspace{1ex}
\begin{IEEEbiography}[{\includegraphics[width=1in,height=1.25in,clip,keepaspectratio]{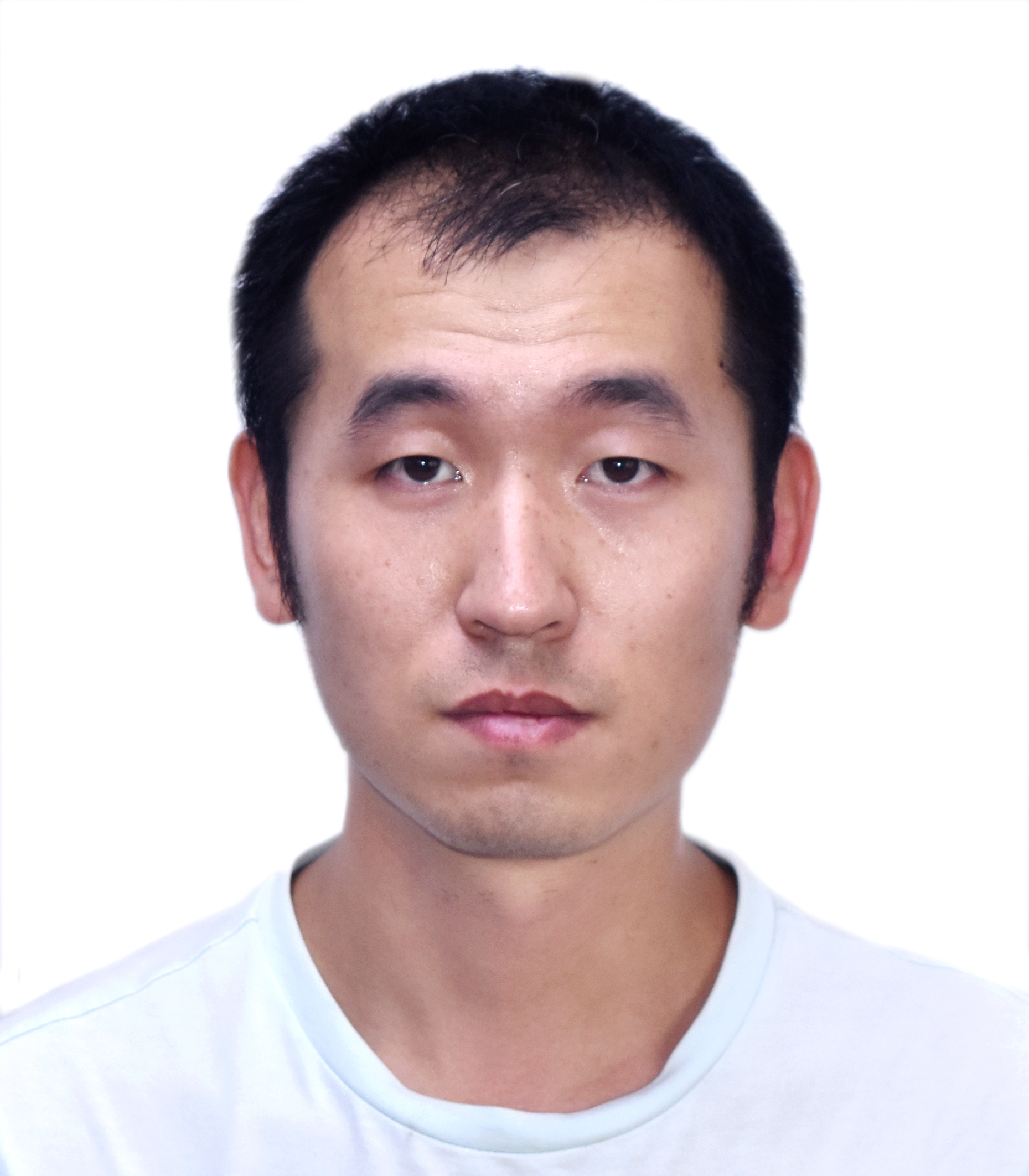}}]{Lingyu Zhang}   joined Baidu in 2012 as a search strategy algorithm research and development engineer. He joined Didi in 2013 and served as senior algorithm engineer, technical director of taxi strategy algorithm direction, and technical expert of strategy model department. Currently a researcher at Didi AI Labs, he used machine learning and big data technology to design and lead the implementation of multiple company-level intelligent system engines during his work at Didi, such as the order distribution system based on combination optimization, and the capacity based on density clustering and global optimization. Scheduling engine, traffic guidance and personalized recommendation engine, "Guess where you are going" personalized destination recommendation system, etc. Participated in the company's dozens of international and domestic core technology innovation patent research and development, application, good at using mathematical modeling, business model abstraction, machine learning, etc. to solve practical business problems. He has won honorary titles such as Beijing Invention and Innovation Patent Gold Award and QCon Star Lecturer, and his research results have been included in top international conferences related to artificial intelligence and data mining such as KDD, SIGIR, AAAI, and CIKM.
\end{IEEEbiography}
\vspace{1ex}
\begin{IEEEbiography}[{\includegraphics[width=1in,height=1.25in,clip,keepaspectratio]{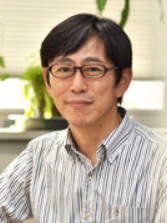}}]{Ryosuke Shibasaki} was born in Fukuoka, Japan. He received his B.S., M.S., and Doctoral degrees in Civil Engineering from The University of Tokyo, Japan, in 1980, 1982, and 1987, respectively. From 1982 to 1988, he was with the Public Works Research Institute, Ministry of Construction. From 1988 to 1991, he was an Associate Professor in the Civil Engineering Department, The University of Tokyo. In  1991, he joined the Institute of Industrial Science, The University of Tokyo. In 1998, he was promoted to Professor in the Center for Spatial Information Science, The University of Tokyo. His research interests cover three-dimensional data acquisition for GIS, conceptual modeling for spatial objects, and agent-based microsimulation in a GIS environment.
\end{IEEEbiography}
\vspace{1ex}
\begin{IEEEbiography}[{\includegraphics[width=1in,height=1.25in,clip,keepaspectratio]{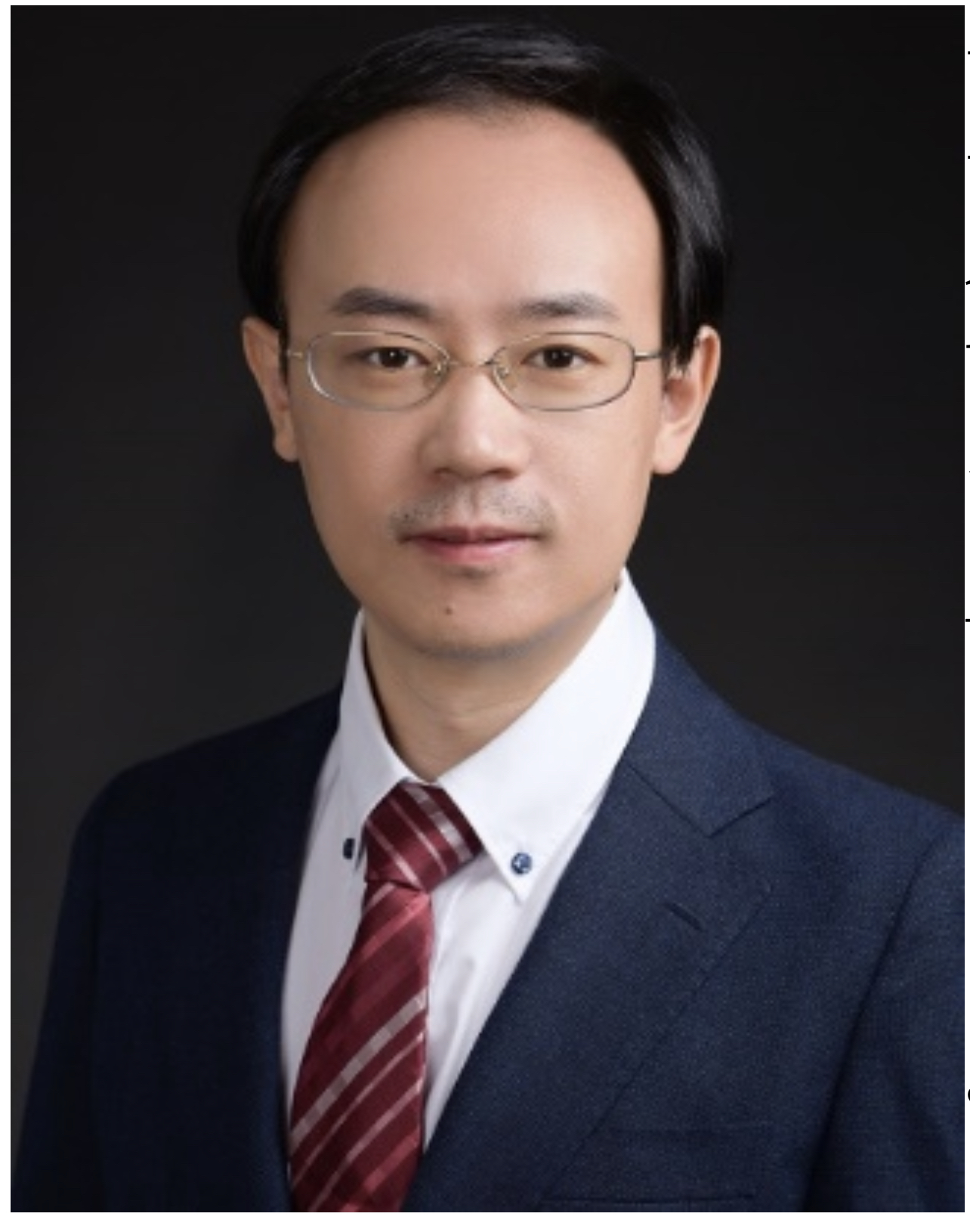}}]{Prof. Xuan Song}  received the Ph.D. degree in signal and information processing from Peking University in 2010. In 2017, he was selected as an Excellent Young Researcher of Japan MEXT. In the past ten years, he led and participated in many important projects as a principal investigator or primary actor in Japan, such as the DIAS/GRENE Grant of MEXT, Japan; Japan/US Big Data and Disaster Project of JST, Japan; Young Scientists Grant and Scientific Research Grant of MEXT, Japan; Research Grant of MLIT, Japan; CORE Project of Microsoft; Grant of JR EAST Company and Hitachi Company, Japan. He served as Associate Editor, Guest Editor, Area Chair, Program Committee Member or reviewer for many famous journals and top-tier conferences, such as IMWUT, IEEE Transactions on Multimedia, WWW Journal, Big Data Journal, ISTC, MIPR, ACM TIST, IEEE TKDE, UbiComp, ICCV, CVPR, ICRA and etc.
\end{IEEEbiography}

\end{document}